\pdfoutput=1

\documentclass[11pt]{article}

\usepackage{acl}

\usepackage{times}
\usepackage{latexsym}

\usepackage[T1]{fontenc}

\usepackage[utf8]{inputenc}

\usepackage{microtype}

\usepackage{inconsolata}

\usepackage{graphicx}

\usepackage{amsmath,amsfonts,bm}

\def\eqref#1{equation~\ref{#1}}

\def\1{\bm{1}}

\DeclareMathAlphabet{\mathsfit}{\encodingdefault}{\sfdefault}{m}{sl}
\SetMathAlphabet{\mathsfit}{bold}{\encodingdefault}{\sfdefault}{bx}{n}

\usepackage{hyperref}
\usepackage{url}
\usepackage{graphicx} 
\usepackage{booktabs}
\usepackage{array}

\usepackage{tabularx}
\usepackage{makecell}
\usepackage{subcaption}
\usepackage{tabularx}

\title{Nova: An Iterative Planning and Search Approach to Enhance Novelty and Diversity of LLM Generated Ideas}

\author{
    Xiang Hu$^{4, *}$,
    Hongyu Fu$^{5, *}$, 
    Jinge Wang$^1$, 
    Yifeng Wang$^6$,
    Zhikun Li$^7$\\
    \textbf{Renjun Xu}$^2$, \textbf{Yu Lu}$^1$, \textbf{Yaochu Jin}$^1$, \textbf{Lili Pan}$^{\dagger,3}$, \textbf{Zhenzhong Lan}$^{\dagger,1}$ \\
    Westlake University$^1$ \;\; Zhejiang University$^2$  \\
    University of Electronic Science and Technology of China$^3$ \\
    China Life R\&D Center$^4$ \;\; Carnegie Mellon University$^5$ \;\; Southeast University$^6$  \;\; University of Oxford$^7$\\
    \texttt{huxiang2022@e-chinalife.com} \; 
    \texttt{hongyuf@andrew.cmu.edu} \;
    \texttt{wangjinge@westlake.edu.cn} \;\\
    \texttt{lilipan@uestc.edu.cn}\; \texttt{lanzhenzhong@westlake.edu.cn}}

\begin{document}
\maketitle
\renewcommand{\thefootnote}{$^*$}
\footnotetext{Equal contribution.}
\renewcommand{\thefootnote}{$^\dagger$}
\footnotetext{Corresponding author.}

\begin{abstract}
Scientific innovation is pivotal for humanity, and harnessing large language models (LLMs) to generate research ideas could transform discovery. However, existing LLMs often produce simplistic and repetitive suggestions due to their limited ability in acquiring external knowledge for innovation. To address this problem, we introduce an enhanced planning and search methodology designed to boost the creative potential of LLM-based systems. Our approach involves an iterative process to purposely plan the retrieval of external knowledge, progressively enriching the idea generation with broader and deeper insights. Validation through automated and human assessments indicates that our framework substantially elevates the quality of generated ideas, particularly in novelty and diversity. The number of unique novel ideas produced by our framework is 3.4 times higher than without it. Moreover, our method outperforms the current state-of-the-art, generating at least 2.5 times more top-rated ideas based on 170 seed papers in a Swiss Tournament evaluation.
\end{abstract}

\section{Introduction}

In recent years, LLMs have demonstrated remarkable progress across various challenging tasks, including solving mathematical problems \citep{romera2024mathematical}, proving mathematical theory \citep{wang2023lego}, and generating code to solve analytical or computational tasks \citep{huang2024mlagentbench}. 
These progresses have opened up new possibilities to utilize LLMs to accelerate research \citep{wang2023scientific}, including generating novel research ideas \citep{si2024llmsgeneratenovelresearch,wang2024scimonscientificinspirationmachines,baek2024researchagentiterativeresearchidea}.


\begin{figure}[ht!]
\centering
\includegraphics[width=0.5\textwidth]{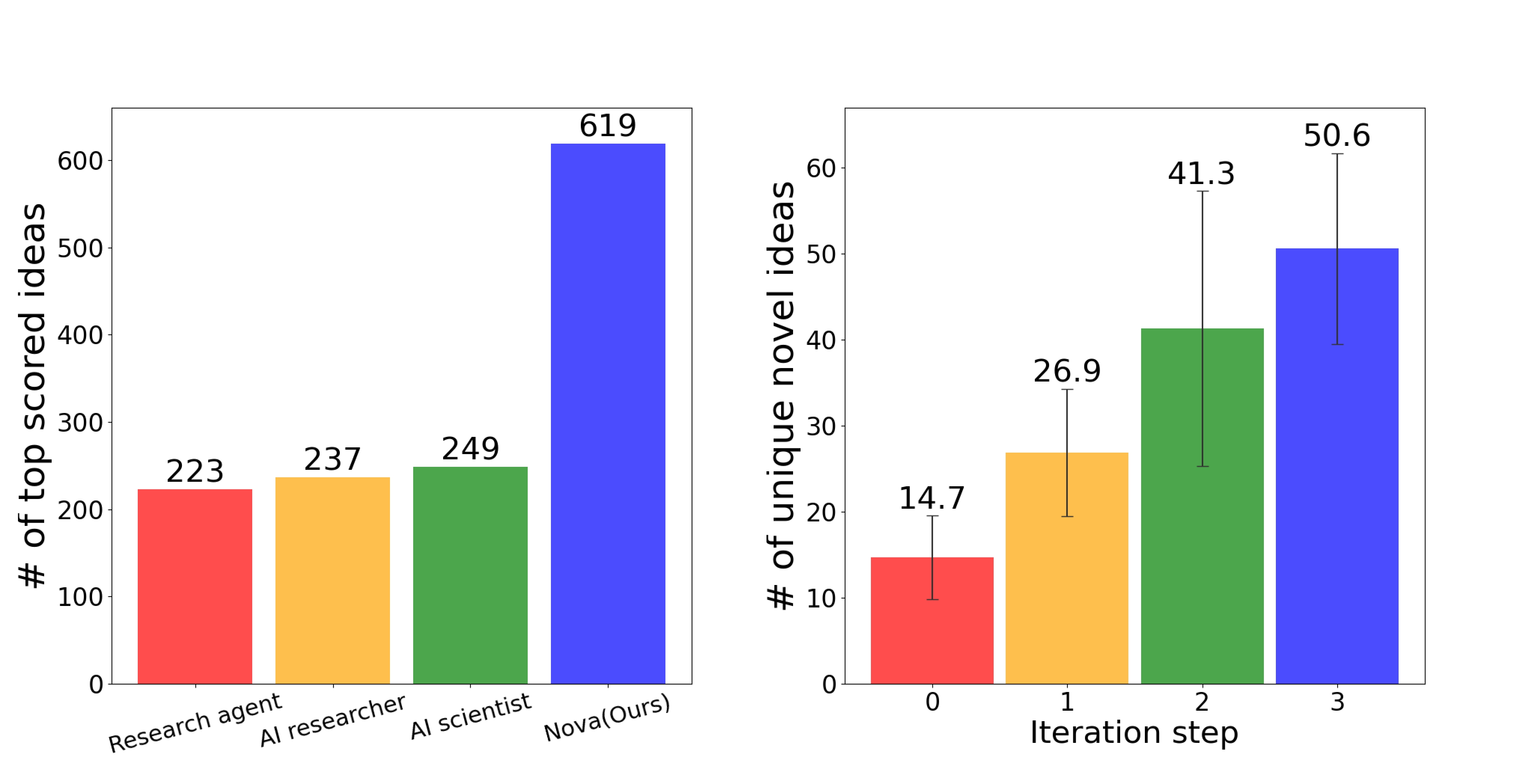}
\caption {\textbf{Nova's Performance.} \textbf{The Left:} Comparison with the state-of-the-arts. Nova significantly outperforms other agents \citep{si2024llmsgeneratenovelresearch, baek2024researchagentiterativeresearchidea, lu2024aiscientistfullyautomated} in generating high-quality ideas (Swiss Tournament Score is 5). \textbf{The Right:} The number of unique novel ideas at each iteration step. The iterative planning framework significantly enhances the generation of unique novel ideas, increasing by 3.4 times from the baseline.}
\label{fig1:five point comparasion and iteration novelty diff}
\end{figure}

Our work is dedicated to addressing the challenge of employing LLMs to produce high-caliber research ideas, with an emphasis on enhancing their novelty and diversity. 
Existing studies \citep{wang2024scimonscientificinspirationmachines, si2024llmsgeneratenovelresearch} tackle this challenge by integrating additional knowledge into the idea generation process. \citet{wang2024scimonscientificinspirationmachines} enrich the process by incorporating co-occurrence entities with existing knowledge, prompting LLMs to generate ideas based on these entities. \citet{si2024llmsgeneratenovelresearch} suggest an iterative approach to retrieve topic-relevant papers through the Semantic Scholar API, utilizing retrieval-augmented generation (RAG) for idea generation. They find that \textit{"LLM-generated ideas are judged as more novel (p < 0.05) than human expert"}. However, they also show that \textit{"LLMs lack diversity in idea generation"}. We argue that this repetitive problem is due to the constrained scope and lack of direction in knowledge acquisition within these methods.

Broadening the search scope, both in terms of breadth and depth, presents a significant challenge. The crux of the issue lies in determining which knowledge to retrieve. Traditional methods of entity and keyword retrieval are not goal-oriented and frequently yield knowledge that is not conducive to fostering innovation.

In order to address the above problem, we introduce an iterative planning framework for LLM-based idea generation that specifically targets the enhancement of the novelty and diversity of the ideas produced. Starting with seed ideas that generated using different scientific discovery methods, our framework undergoes multiple iterations of planning and searching. In each iteration, the model is tasked with devising a search plan aimed at identifying papers that will enhance the novelty and diversity of the current set of ideas.

As depicted in Fig. \ref{fig1:five point comparasion and iteration novelty diff},the proposed iterative planning framework significantly enhances the quality of ideas generated from recent 170 LLM-related papers (from top conferences like ACL, ICLR, and CVPR). The number of high-quality ideas (as measured by the Swiss Tournament Score \cite{si2024llmsgeneratenovelresearch}) is at least 2.5 times greater than those produced by other state-of-the-art methods. Moreover, the number of unique novel ideas generated by our iterative planning framework is 3.4 times higher compared to approaches that do not incorporate such a framework.

\section{Related work}\label{related work}
\label{gen_inst}

\subsection{LLM-based Scientific Innovation}

In the past year, several studies on LLM-based scientific innovation \citep{yang2024largelanguagemodelsautomated, baek2024researchagentiterativeresearchidea, lu2024aiscientistfullyautomated, wang2024scimonscientificinspirationmachines, gu2024generationhumanexpertevaluationinteresting, li2024mlrcopilotautonomousmachinelearning} have been proposed, garnering significant attention from the LLM community.
Among these studies, \citet{baek2024researchagentiterativeresearchidea} introduces a research agent that utilizes an external knowledge graph for co-occurrence entity search and integrates retrieved entities into idea generation of LLMs.
To avoid generating similar ideas, 
\citet{lu2024aiscientistfullyautomated} treat past generated ideas as negative examples and instruct the LLM on what constitutes a negative example. To explore more external knowledge for innovation, some other works \citep{wang2024scimonscientificinspirationmachines, gu2024generationhumanexpertevaluationinteresting} propose prompting the LLM to generate ideas integrated with external knowledge, such as retrieved external entities or problem-solution pairs. 

Concurrent with our research, \citet{si2024llmsgeneratenovelresearch} introduce AI-Researcher, which, for the first time, demonstrates that LLMs can generate ideas deemed more novel than those written by human experts.
In addition, they point out that using LLMs to directly evaluate different dimensions of scientific ideas is unreliable and propose an idea ranking method based on pairwise comparison, achieving an accuracy of 71.4\% in distinguishing accepted and rejected submissions on real ICLR 2024 data.

Although effective, the above approach often generates repetitive ideas \cite{si2024llmsgeneratenovelresearch} due to the lack of direction in acquiring new knowledge. In contrast, our method provides a plan for searching for new knowledge and suffers less from the repetitive problem. 

\subsection{Reasoning and Planning}

Reasoning has been proven to be an effective technique for enhancing the problem-solving capabilities of LLMs, and several studies have been conducted to further promote LLMs' reasoning abilities. \citet{wei2022chainofthought} propose chain of thought (CoT), which involves guiding LLMs to solve complex problems by generating a step-by-step reasoning process. Later, \citet{wang2022selfconsistency} improve CoT by sampling and comparing diverse reasoning pathways to enhance the consistency of the reasoning process.
To solve problems harder than the exemplars shown in prompts, \citet{zhou2022leasttomost} propose to break down the complex problem into a series of simpler subproblems and then solve them in sequence.
Generalizing from Chain of Thought (CoT), \citet{yao2023treethoughts} propose the Tree of Thought (ToT) framework, enabling LLMs to explore multiple reasoning paths and conduct self-evaluations when determining the next action.
To enable more effective exploration of the solution space, \citet{xie2024montecarlo} enhance the reasoning capabilities of LLMs by introducing Monte Carlo Tree Search (MCTS) with iterative preference learning.

 \begin{figure*}[t]
\centering
\includegraphics[width=\textwidth]{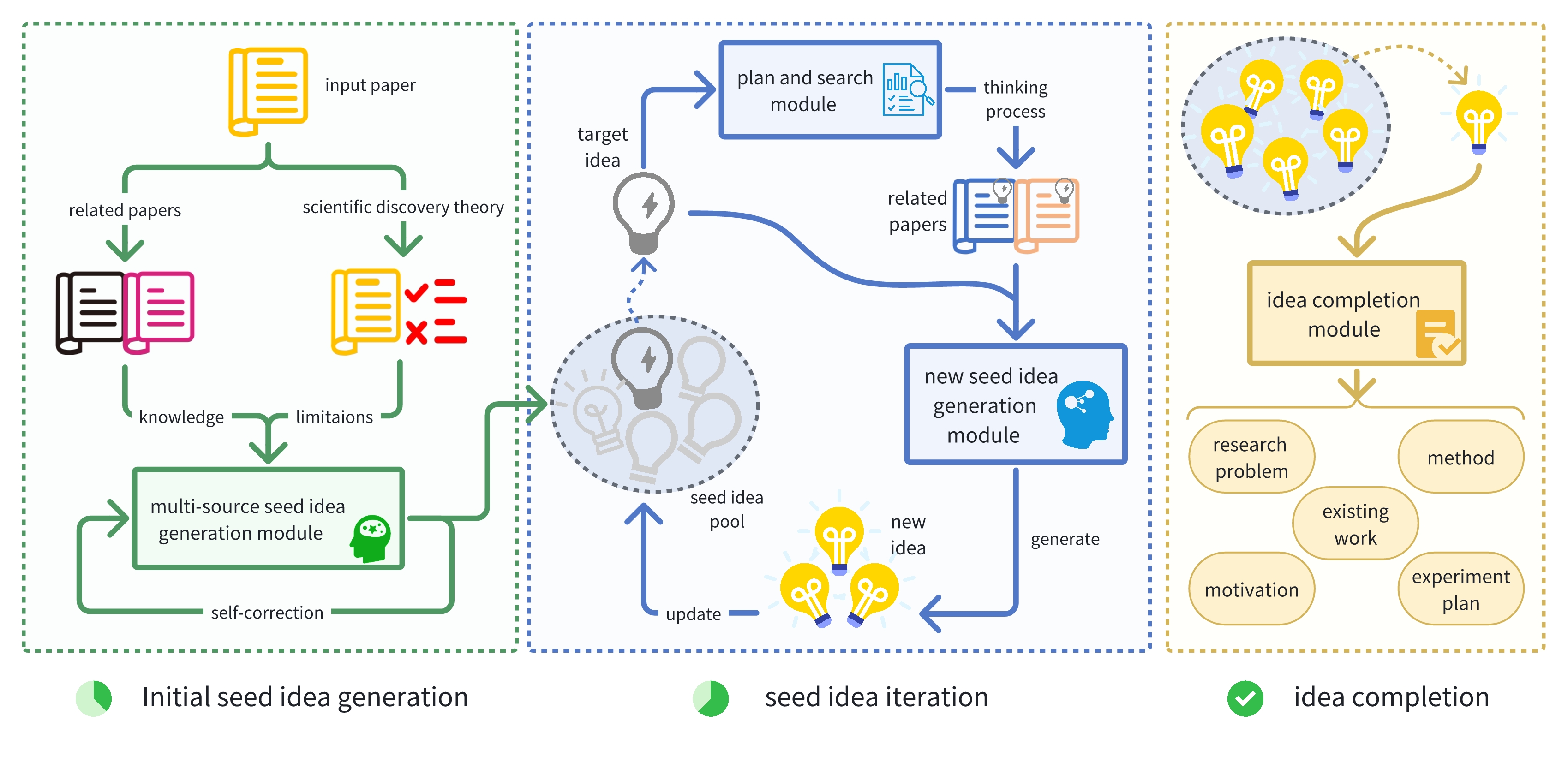}
\caption {\textbf{Nova Pipeline.} The Pipeline includes initial seed idea generation, seed idea iteration, and idea completion. Upon receiving an input paper (\emph{i.e.}, seed paper), the LLM is prompted to generate initial seed ideas by utilizing related papers (including recent publications) and scientific discovery methods. After that, the generated ideas are revised according to the new knowledge acquired according to iterative planning and search. Finally, each idea is expanded with more detailed methods. }
\label{fig5:iterate idea}
\end{figure*}

These methods significantly enhance the reasoning capabilities of LLMs; however, they seldom consider interacting with the external environment.
To address this limitation, \citet{trivedi2022interleaving} integrate CoT with knowledge retrieval, interleaving reasoning with searching to acquire additional external knowledge for knowledge-intensive question answering.
\citet{yao2022react} propose a ReAct paradigm combining reasoning and acting for solving language reasoning and decision-making tasks. It creates and adjusts high-level plans for acting while also interacting with the external environments to incorporate additional information into reasoning. Later, \citet{aksitov2023restmeetsreact} develop a ReAct-style LLM agent to reason and act upon external knowledge, using self-critique for self-improvement.
By integrating reasoning and acting, these methods achieve more dynamic and contextually aware problem-solving based on both internal knowledge and external knowledge.

The reasoning capabilities of LLMs can also be applied to planning, such as generating plausible goal-driven action plans that can be enacted in interactive, embodied environments \citep{huang2022language}. Additionally, plan-to-solve prompting \citep{wang2023plan} can generate a plan that divides complex reasoning tasks into subtasks, enabling LLMs to execute each subtask according to the outlined plan.

Our work marks the inaugural integration of planning methodologies into the complex domain of research tasks.

\begin{figure*}[h]
\centering
\includegraphics[scale=0.16]{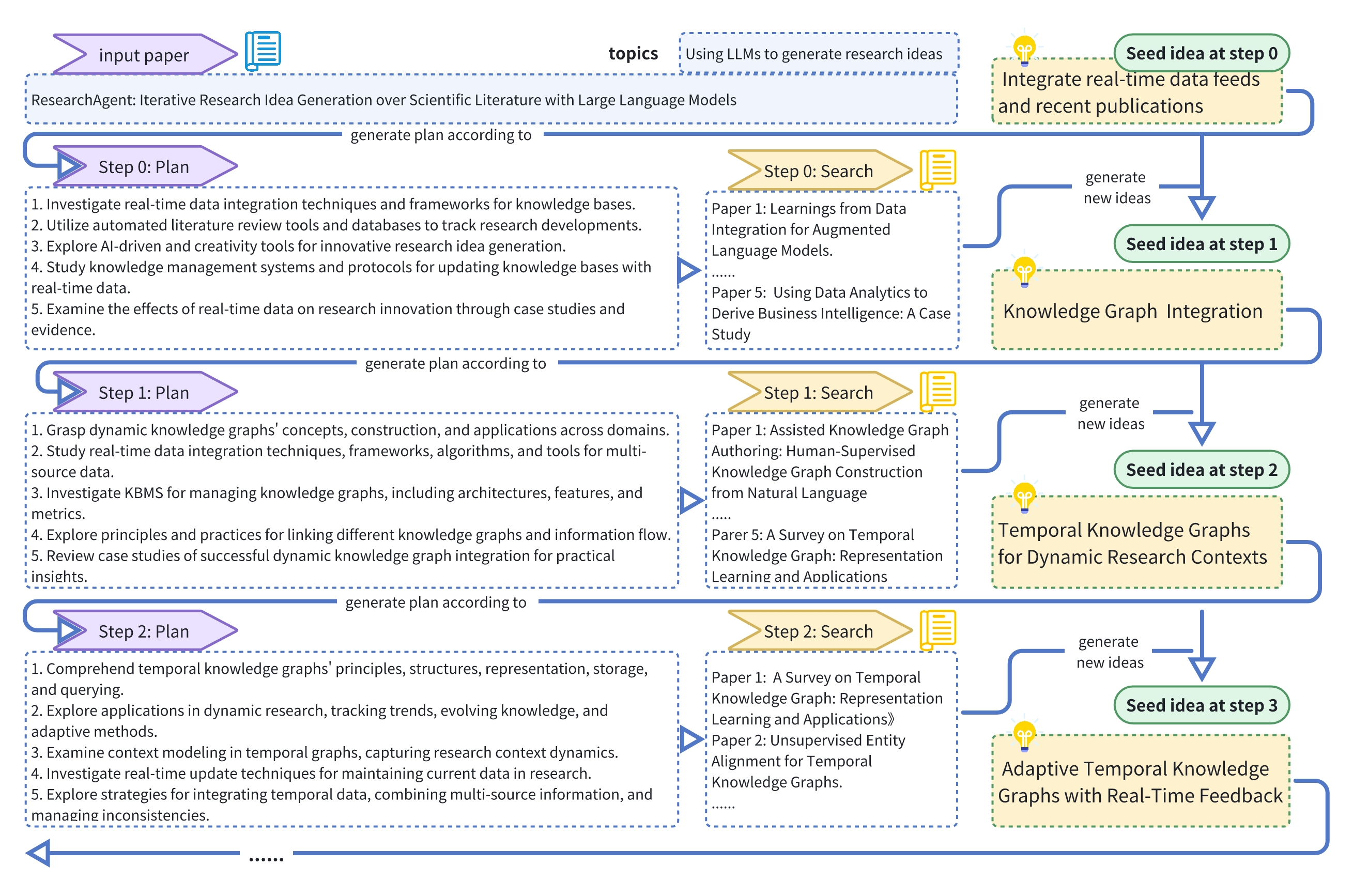}
\caption {\textbf{Example of Planning-Driven Iterative Seed Idea Generation Process.} This example highlights the planning-driven iterative seed idea iteration process. Starting from an initial concept, a detailed plan is formulated to guide the search for relevant literature and acquire up-to-date knowledge.}
\label{fig5.1:plan_iteration_example}
\end{figure*}

\section{Nova Pipeline}

The pipeline for Nova is illustrated in Fig. \ref{fig5:iterate idea}. Our pipeline streamlines the research process through three stages: initial idea generation, iterative refinement, and detailed completion. It begins with an input paper, which the LLM uses to generate initial ideas by drawing on related literature and scientific discovery techniques. These ideas are then enhanced through iterative planning and search, incorporating new insights. The final step involves detailing the ideas. An example of the whole process is in Fig. \ref{fig5.1:plan_iteration_example}.

\begin{table*}[ht]
\centering
\begin{tabular}{p{15cm}}
\toprule
\textbf{Prompt} \\
\midrule
\textbf{Role:} You are an expert researcher in AI. Your goal is to propose some innovative and valuable research ideas based on the target paper. \\
\textbf{Skill:} Generate subsequent exploration ideas according to the following steps: \\
Understanding of the target paper and related papers is essential: \\
- The target paper is the primary research study you aim to enhance or build upon through future research, serving as the central source and focus for identifying and developing the specific research idea. \\
- The referenced papers are studies that the target paper has cited, indicating their direct relevance and connection to the primary research topic you are focusing on, and providing additional context and ideas that are essential for understanding and expanding upon the target paper.  \\
\textbf{Step 1:} Combine target paper and referenced paper to answer the following information: \\
1. What are the tasks, methods, and main innovations of the current paper? \\
2. What are the weaknesses and limitations of the current paper? \\
\textbf{Step 2:} Propose some valuable and new research ideas. \\
\textbf{Output Format:} \{qa\_info\_with\_idea\_json\_format\} \\
\textbf{Requirements:} ... \\
\textbf{Input:} ...\\
\textbf{Output:} ... \\

\bottomrule
\end{tabular}
\caption{Prompt for initial seed idea generation.}
\label{tab:initial ideal generation prompt}
\end{table*}

\subsection{Initial Seed Idea Generation}\label{seed idea generate}

To produce high-quality ideas, we design a multi-source seed idea generation module that initiates with diverse and novel concepts. This module activates the LLM to generate ideas using related literature and scientific discovery techniques upon receiving an input paper. The prompt for initial idea generation is in Tab. \ref{tab:initial ideal generation prompt} (details in Tab. \ref{tab:table2}) and an example of an initial seed idea is in Tab. \ref{tab: seed idea example}.

\begin{table*}[ht]
\centering
\begin{tabular}{p{15cm}}
\toprule
\textbf{Thinking:} The target paper’s reliance on existing literature may limit the generation of truly novel ideas. By incorporating real-time data sources, such as ongoing research developments or recent publications, the ResearchAgent can
generate more innovative and timely research ideas.\\ \midrule
\textbf{Idea:} Incorporate multi-modal data sources, including experimental data, patents, and industry reports, into the ResearchAgent’s knowledge base to generate more comprehensive and interdisciplinary research ideas.\\ \midrule
\textbf{Keywords:} multi-modal data sources, experimental data, interdisciplinary research ideas\\
\bottomrule
\end{tabular}
\caption{Seed Idea Example.}
\label{tab: seed idea example}
\end{table*}

To enrich the knowledge base with the most current insights, we utilize the input paper's references and have designed a knowledge tracking module. This module addresses the shortcomings of previous approaches by monitoring the latest publications. We pinpoint influential recent papers based on user engagement metrics such as likes, comments, and reposts across social media, forums, and GitHub. Furthermore, we harness LLMs to distill summaries of prevailing research trends from these papers, extracting valuable knowledge to enrich our target innovation efforts.

To further increase the diversity of the generated ideas, we employ 10 fundamental scientific discovery methods to guide LLMs in generating innovative ideas from an input paper and its associated literature. Drawing on Kuhn's paradigm \citep{kuhn1997structure} of scientific discovery, these methods help identify new research problems, such as analyzing anomalies in existing approaches. An example theory of scientific discovery can be found in Tab. \ref{tab:theory_example} and all theories (Appendix Tab. \ref{tab:discovery_method_1} and \ref{tab:discovery_method_2}.)

\begin{table*}[ht]
\centering
\begin{tabular}{p{15cm}}
\toprule

\textbf{Define New Scientific Problems} \\
\textbf{Theoretical Basis:} Kuhn's paradigm theory, Laudan's problem-solving model, Nichols's problem-generation theory. \\
\textbf{Method:} Identify anomalies in existing theories; explore theoretical boundaries and scope of application; integrate interdisciplinary knowledge and discover new problems; re-examine neglected historical problems. \\ \midrule

\end{tabular}
\caption{An example theory of scientific discovery}
\label{tab:theory_example}
\end{table*}

To mimic human intuition, we tap into the LLMs' internal knowledge to craft initial seed ideas. This involves prompting the LLM to assess the shortcomings of the input paper and related works, thereby sparking the creation of fresh ideas.

To prevent hallucination and improve the logicality of generated initial seed ideas, we also utilize self-correction mechanics: self-check \citep{miao2023selfcheckusingllmszeroshot}, self-critique \citep{gou2024criticlargelanguagemodels}, and reflection \citep{shinn2023reflexionlanguageagentsverbal}. These methods partly guarantee that the generated seed ideas are logical and reasonable. In the end, we generate 15 seed ideas for each input paper.

\subsection{Iterative Planning and Search for Seed Idea Improvement}\label{iterate idea}

Once an initial seed idea pool is generated, we start to iteratively planning and search new knowledge according to the see idea and generate new idea using the acquired new knowledge.

\subsubsection{Planning and Search}

In planning and search step, we guide the LLM to identify key fields for comprehensive and novel knowledge acquisition to enhance further research and idea generation based on the given ideas. 

This approach, demonstrated through an in-context learning example, leverages the LLM's internal knowledge to determine useful knowledge for new ideas, surpassing traditional entity or keyword-based retrieval methods.

\noindent \textbf{New Seed Idea Generation.}  Once new knowledge is acquired, the new seed idea is generated based on the retrieved papers, the initial seed idea, and the given input paper. For each idea, our models generate 10 new seed ideas and then use self-reflection to cut the number down to 3.  

In each iteration, the old seed ideas are replaced with the newly generated seed ideas. This allows our agent to dive deeper, largely expanding the scope of search. Therefore, in each iteration, we generate 3 times more seed ideas.

\subsection{Output Idea Generation} 

After finishing $T$ step iteration, we have a final seed idea pool.
We then expand the seed idea into the initial proposal and final proposal as in \cite{si2024llmsgeneratenovelresearch}. Specifically, given an input paper and its corresponding seed idea, we ask LLM to decompose the idea into several sub-modules and utilize LLMs to design these sub-modules separately in a more detailed way(details in Tab. \ref{tab:table20} in Appendix).  An example of an initial proposal and final proposal are in Tab. \ref{tab: initial proposal template} and Tab. \ref{tab: Final proposal template}, separately.

\begin{table*}[ht]
\centering
\begin{tabular}{p{15cm}}
\toprule
\textbf{Problem:} State the problem statement, which should be closely related to the idea description and something that large language models cannot solve well yet.\\ \midrule
\textbf{Existing Methods:} Mention some existing benchmarks and baseline methods if there are any.\\ \midrule
\textbf{Motivation:} Explain the inspiration of the proposed method and why it would work well.\\ \midrule
\textbf{Proposed Method:} Propose your new method and describe it in detail. The proposed method should be maximally different from all existing work and baselines, and be more advanced and effective than the baselines. You should be as creative as possible in proposing new methods; we love unhinged ideas that sound crazy. This should be the most detailed section of the proposal.\\ \midrule
\textbf{Experiment Plan:} Specify the experiment steps, baselines, and evaluation metrics.\\
\bottomrule
\end{tabular}
\caption{Initial Proposal Template (follow \citet{si2024llmsgeneratenovelresearch}).}
\label{tab: initial proposal template}
\end{table*}

\begin{table*}[ht]
\centering
\begin{tabular}{p{15cm}}
\toprule
\textbf{Title:} A concise statement of the main research question to be used as the paper title.\\ \midrule
\textbf{Problem Statement:} Clearly define the problem your research intends to address. Explain clearly why this problem is interesting and important.\\ \midrule
\textbf{Motivation:} Explain why existing methods are not good enough to solve the problem, and explain the inspiration behind the new proposed method. You should also motivate why the proposed method would work better than existing baselines on the problem.\\ \midrule
\textbf{Proposed Method:} Explain how the proposed method works, describe all the essential steps.\\ \midrule
\textbf{Step-by-Step Experiment Plan:} Break down every single step of the experiments, make sure every step is executable. Cover all essential details such as the datasets, models, and metrics to be used. If the project involves prompting, give some example prompts for each step.\\
\bottomrule
\end{tabular}
\caption{Final Proposal Template (follow \citet{si2024llmsgeneratenovelresearch}).}
\label{tab: Final proposal template}
\end{table*}

\section{Experiment}\label{experiment}
\label{others}
To validate our proposed iterative planning framework, we perform comprehensive comparisons with state-of-the-art research idea generation methods and conduct an ablation study.

\subsection{Experimental Setup}
\textbf{Data.} Our dataset is constructed by collecting high-quality papers from top conferences. 
The initial corpus comprised 7,805 papers from  CVPR 2024, ACL 2024, and ICLR 2024. 
The keywords related to ``LLM" are used to filter the initial corpus down to about 2,000 papers. The minimum citation number is used to further cut down the number to 153, with citation thresholds set at 20 for ICLR 2024 and 10 for CVPR 2024 and ACL 2024. We add additional 17 papers from Hugging Face Daily Papers according to their user ratings. At the end, the dataset consists of 170 papers, each of which is used to generate 100 ideas for subsequent evaluation.

\noindent\textbf{Baseline.}  
To compare our proposed approach with the state-of-the-art approaches, we choose three leading approaches as the baselines, including AI-Researcher \citep{si2024llmsgeneratenovelresearch},  AI-Scientist \citep{lu2024aiscientistfullyautomated} and Research-Agent \citep{baek2024researchagentiterativeresearchidea}. 
For AI-researcher and AI-Scientist, we run the original code on our seed papers. For Research-Agent, we also build a knowledge graph based on the description of the paper.

\noindent\textbf{Automatic Evaluation.} In our automatic evaluation, we mainly focus on overall quality evaluation and also concern with the novelty and diversity.

\noindent\textbf{1. Quality.} Following \citet{si2024llmsgeneratenovelresearch}, we employ the Swiss System Tournament \footnote{https://en.wikipedia.org/wiki/Swiss-system\_tournament} with Claude-3.5-Sonnet zero-shot ranker to evaluate the quality of ideas. 
The ranker makes pairwise comparisons to determine which idea is better. For each idea, there are 5 rounds of comparison, each winning comparison gets 1 score. 
Such a quality evaluation method has been shown to be better than direct comparison \citep{lu2024aiscientistfullyautomated}.

\noindent\textbf{2. Novelty.} Following \citet{baek2024researchagentiterativeresearchidea}, we use LLMs to judge whether a generated idea is novel by checking the top 10 most relevant papers \footnote{We use the same paper retriever as in the idea generation phase, with data sourced from arXiv. The time range spans from January 1, 2022, to August 2024, and the categories include literature related to AI, NLP, and CV.} and if no paper is identified as containing a similar idea, it is considered novel. We use embedding using the all-MiniLM-L6-v2 model\footnote{https://huggingface.co/sentence-transformers/all-MiniLM-L6-v2} and if the cosine similarity threshold is larger than 0.3, we say that they are similar \citep{si2024llmsgeneratenovelresearch}.


\noindent\textbf{3. Diversity.} Similar to \citet{si2024llmsgeneratenovelresearch}, we use the proportion of unique ideas to measure generation diversity. To be specific, we use the same similarity measurement as in the novelty measurement and the duplication threshold is set to be 0.8.

\begin{figure}[t]
\centering
\includegraphics[scale=0.25]{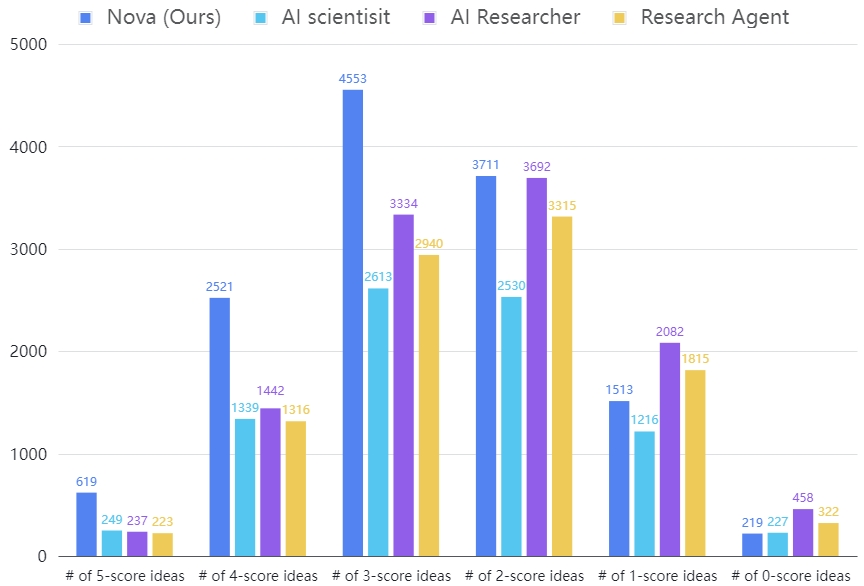}
\caption {Score distribution of different methods in Swiss Tournament. The results indicate that Nova not only generates more unique ideas but also produces a greater proportion of high-quality ideas. 619 and 2521 ideas generated by Nova are scored at 4 and 5, significantly surpassing the baseline methods.}
\label{fig7:swiss score distribution}
\end{figure}

In the automatic evaluation, we utilize Nova alongside three baseline methods to generate 400 ideas from a given input paper. Each method generates 100 ideas separately. The iteration step for Nova is set to 3, and the initial number of seed ideas is 15. After 3 iterations, we get 405 initial seed ideas, we then cluster the ideas using k-means clustering into 100 clusters and use the cluster center to generate the 100 final ideas. The implementation details for the baseline methods remain consistent with those in the original work.

\noindent\textbf{Human Evaluation.} To validate the effectiveness of our automatic evaluation, we have an additional human evaluation.
Our goal is to assess how well our automatic evaluation aligns with human expert evaluations. 
We recruit a panel of 10 experts, all holding a PhD degree or professorship in natural language processing, machine learning, or computer vision, doing research in LLMs-related fields.
These experts evaluated ideas based on novelty and overall quality (including feasibility and effectiveness).

We select five ideas generated by each agent based on the same input paper. These ideas correspond to the $1^{st}$, $25^{th}$, $50^{th}$, $75^{th}$, and $100^{th}$ percentiles of the automatic evaluation, resulting in a total of $20$ ideas per topic. This process is repeated for $20$ times. Each expert reviews four groups, ensuring that at least two independent experts evaluate each idea. The final score for each idea is averaged across all ratings from different experts.

We compare the distributions of expert evaluations against those in automatic evaluation. Specifically, we track which methods produce the top 20 percent of ideas, as ranked by the experts. This helps determine which methods outperform the others. Moreover, this approach reveals whether the model evaluation aligns with human evaluation. 

\subsection{Experimental Results}

\subsubsection{Automatic Evaluation Results}
\begin{figure}[t]
\centering
\includegraphics[width=\columnwidth]{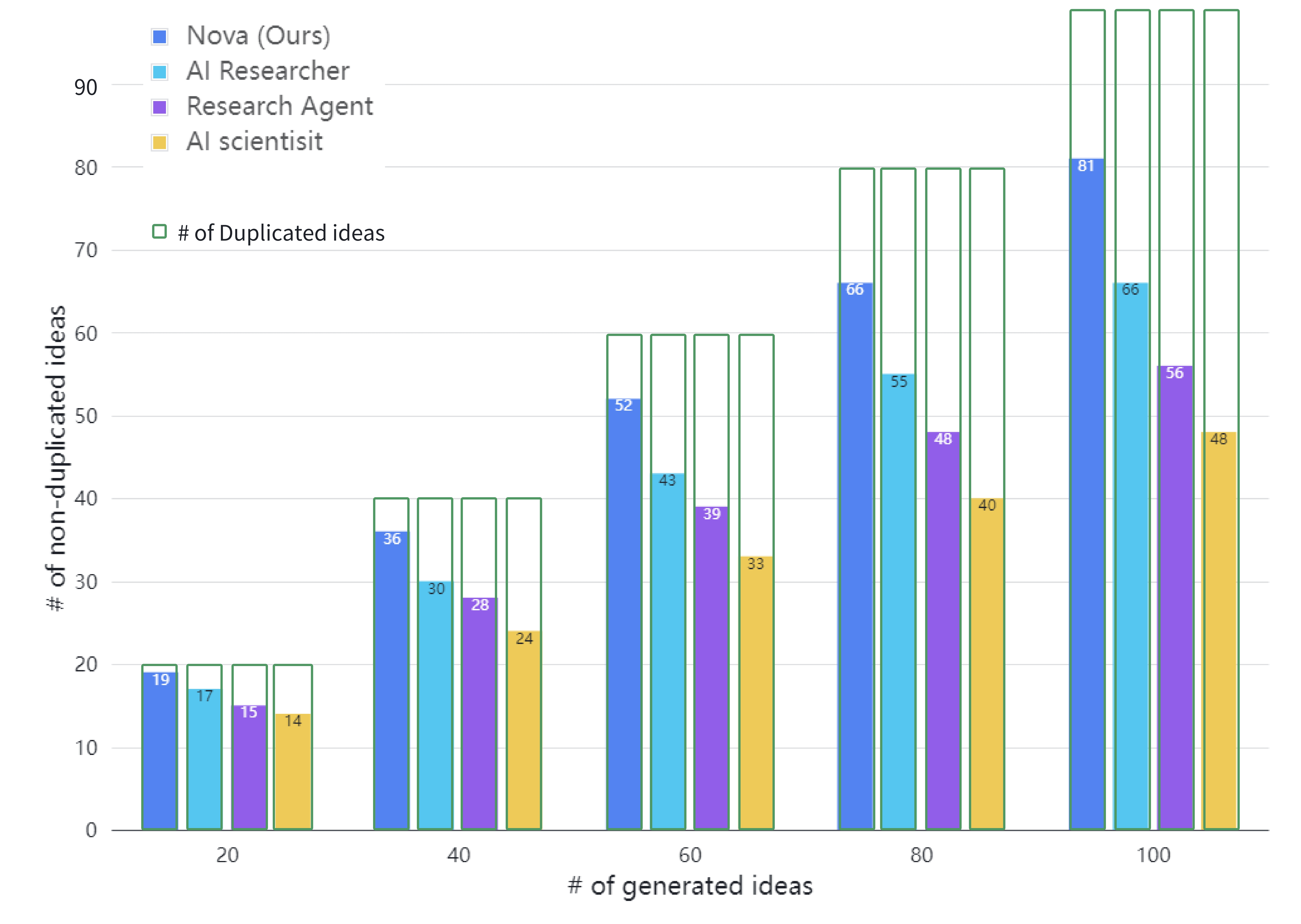}
\caption {Non-Duplicate Percentage Comparison.} 
\label{fig1:diversity vs baseline}
\end{figure}

The Swiss Tournament score comparison are shown in Fig. \ref{fig7:swiss score distribution}.
The novelty and diversity comparison are shown in Fig. \ref{fig1:diversity vs baseline}.

Clearly, Nova achieves a significantly higher Swiss score. 619 and 2521 of the ideas generated by Nova are scored at 4 and 5, significantly surpassing the performance of other agents.
By incorporating iterative planning and search for external knowledge retrieval, Nova engages in more effective exploration for innovation. This may significantly enhance the novelty of the generated ideas. Since novelty is often the most important factor in evaluating idea quality, Nova consistently better than other state-of-the-art methods.


Fig. \ref{fig1:diversity vs baseline} shows that Nova generates significantly more diverse ideas. 
As the number of generated ideas increases, Nova can continuously generate new ideas through iterative planning and search. 
In Non-Duplicate Percentage, Nova significantly outperforms others, with over 80\% of the ideas being unique.

\subsubsection{Human Evaluation Results}

\begin{figure}
    \centering
    \includegraphics[width=\columnwidth]{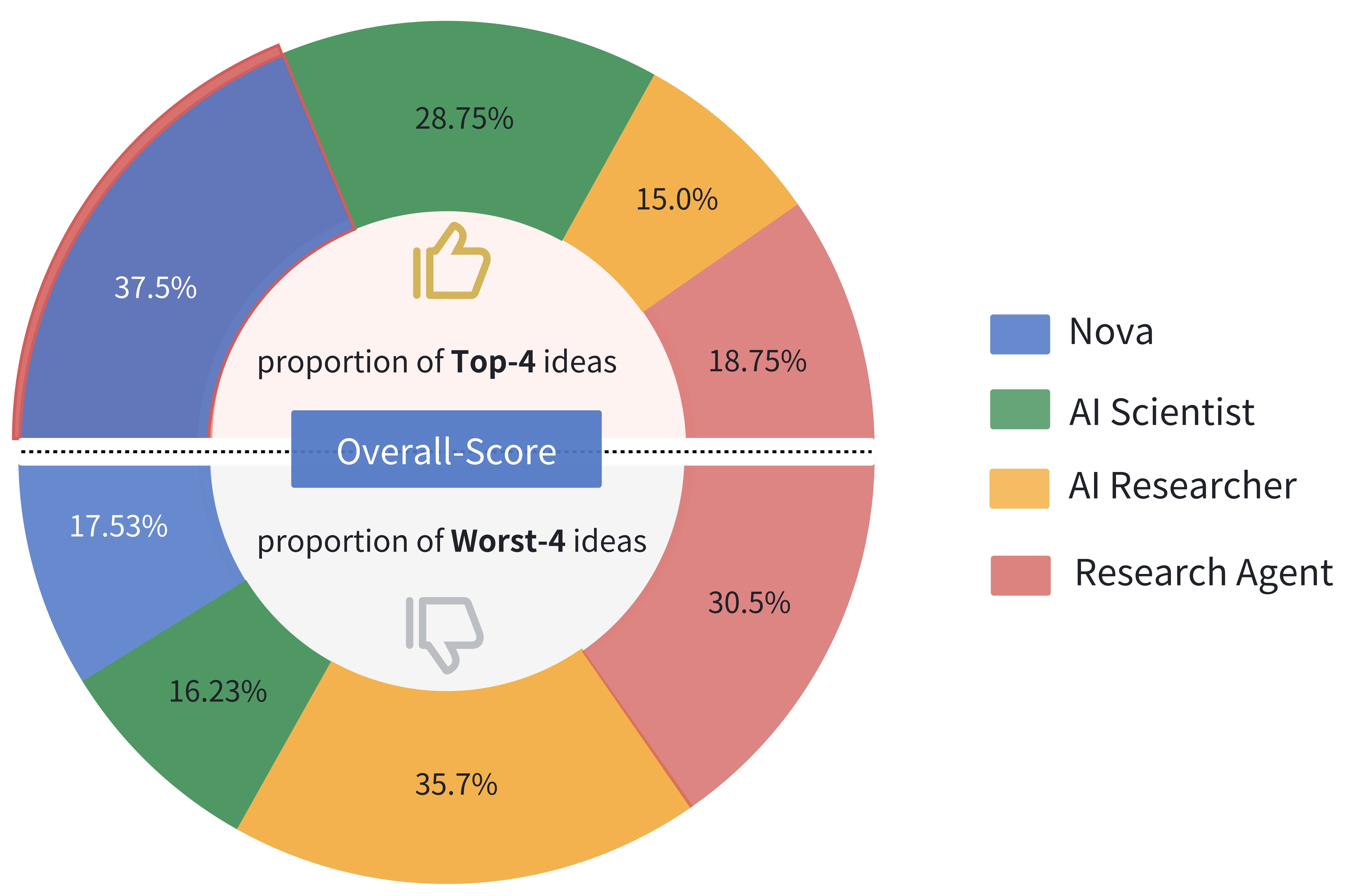}
    \caption{
     Human Evaluation for Overall Quality. 
    }
    \label{figonline_overall}
\end{figure}
\begin{figure}
    \centering
    \includegraphics[width=\columnwidth]{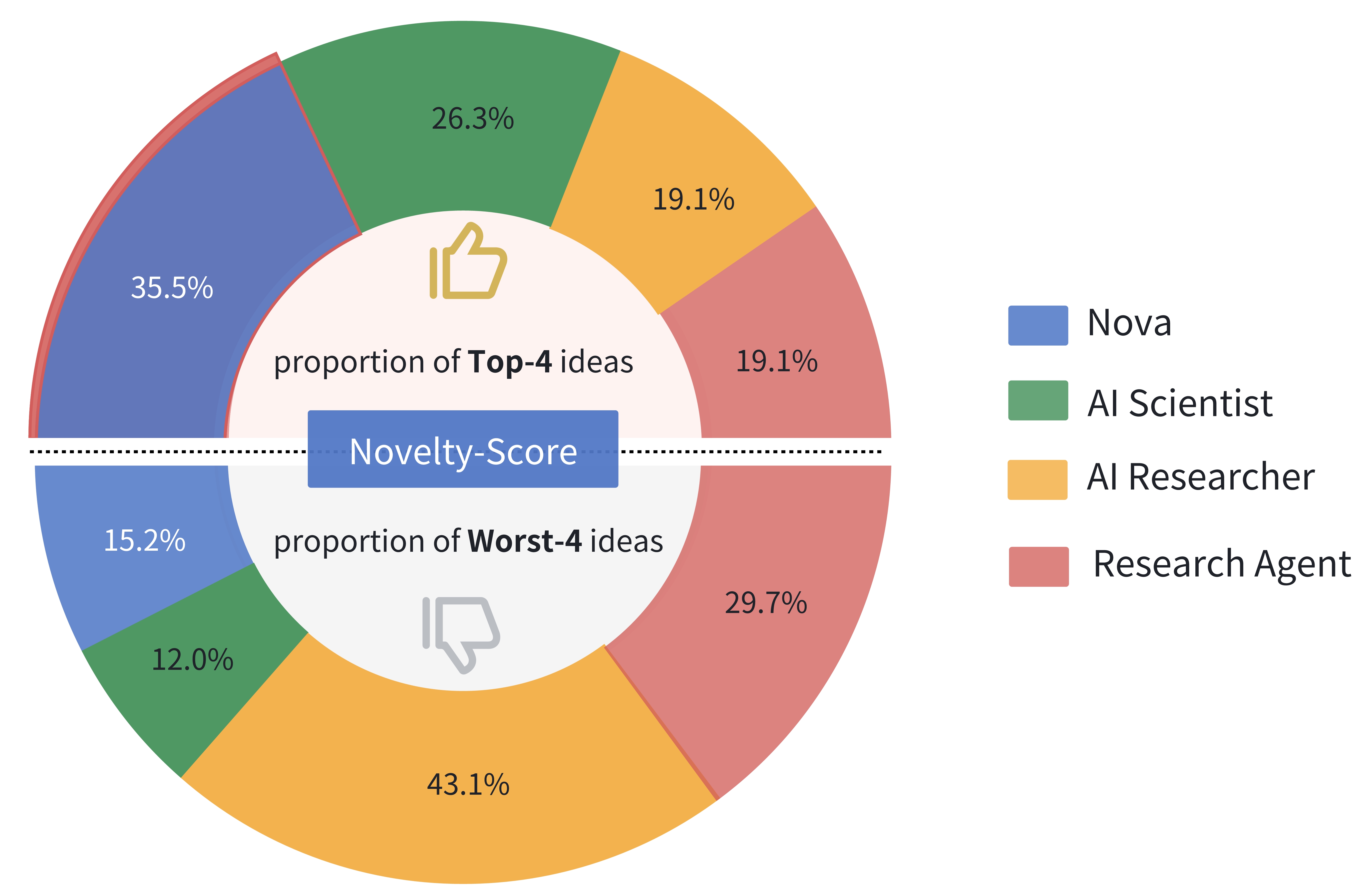}
    \caption{Human Evaluation for Novelty. 
    }
    \label{figonline_novelty}
\end{figure}

In our human evaluation, Nova achieves the highest scores for both overall quality and novelty. As shown in Fig. \ref{figonline_overall}, Nova contributes $37.5\%$ of the top 4 ideas, the highest among the four methods. Additionally, Nova has a notably low percentage of the worst 4 ideas, accounting for only $17.53\%$ in terms of overall quality. 
In Fig. \ref{figonline_novelty}, a similar pattern is observed in novelty evaluation.


Our human and automated evaluations show strong consistency in distinguishing between the top-rated and worst-rated ideas. By comparing the distribution of top-rated ideas in both human and automated evaluations (Fig. \ref{fig7:swiss score distribution} and \ref{figonline_overall}), it is evident that human reviewers and the LLM evaluate the performance of the four methods in a similar pattern. In both human and automatic evaluations, our method generates the highest proportion of top-rated ideas, followed by AI-Scientist, ResearchAgent, and finally AI-Researcher. This indicates that our automatic review mechanism effectively captures human reviewers' true preferences.


\subsection{Ablation Study}

To assess the effectiveness of planning and search in Nova, we conduct comparisons by gradually removing planning and retrieval components. All methods retrieve the same number of papers, specifically $K=5$. Both retrieval and planning are found to significantly enhance the generation of unique and novel ideas. When planning is excluded, the number of unique ideas at step 3 ($44.1$) no longer increases compared to step 2 ($42.4$). This suggests that without planning, relying solely on retrieval based on seed ideas limits access to valuable external knowledge for innovation. This limitation may arise from the restricted scope of search when planning is absent.
Obviously, when planning and retrieval are both removed, the number of unique novel ideas increases slightly at step 2 (from $25.3$ to $30.6$) and stagnates at step 3 (from $30.6$ to $31.35$), due to no external knowledge being introduced.
\begin{figure}[t]
    \centering
    \includegraphics[width=1.0\linewidth]{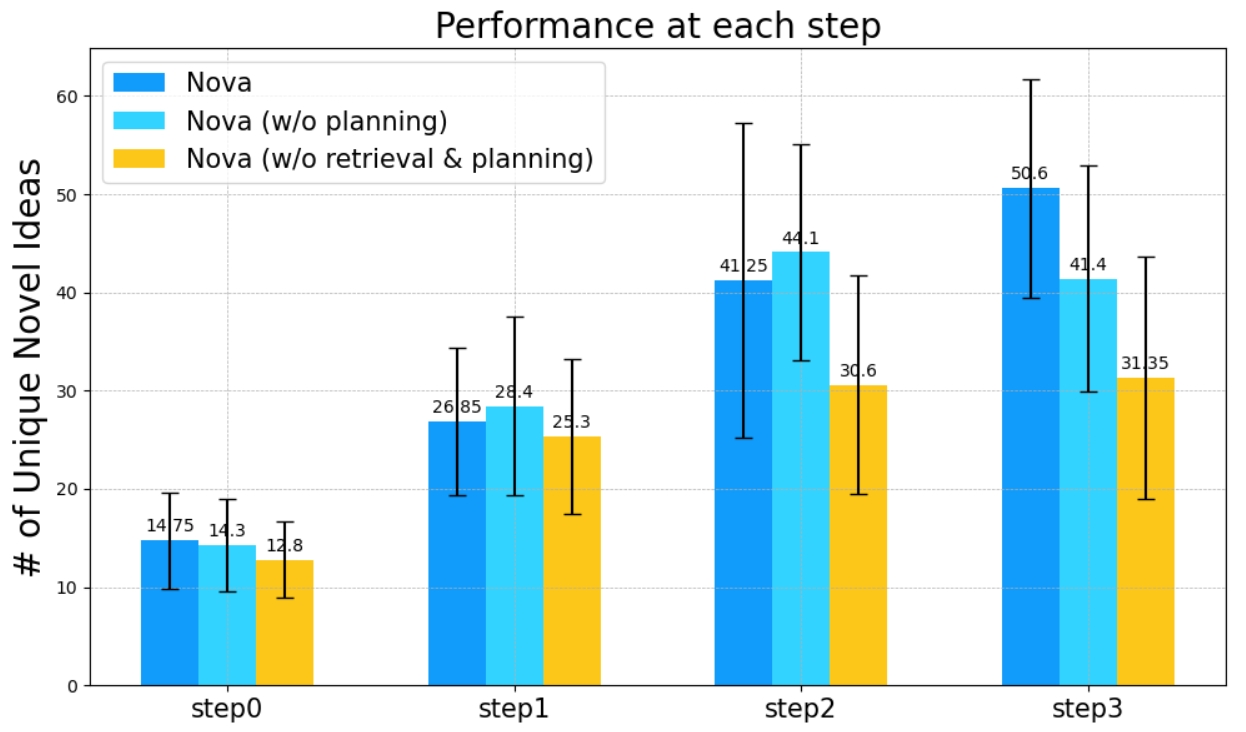}
    \caption{Ablation studies for Nova. We can find that both retrieval and planning significantly enhance the generation of unique novel ideas.}
    \label{fig:enter-label}
\end{figure}

\section{Conclusion}
In this paper, we propose an LLM-based scientific innovation method, Nova, which introduces iterative planning and search to retrieve external knowledge for innovation. Nova leverages the internal knowledge of LLMs to generate search plans for external knowledge retrieval, significantly enhancing the effectiveness of the retrieval process.
The ablation study demonstrates the effect of the iterative planning and search framework on promoting the novelty of generating ideas.
The automatic and human evaluations show that Nova significantly and consistently outperforms state-of-the-art scientific innovation methods.
In the future, we will explore incorporating a reward function into our iterative planning framework to further enhance external knowledge retrieval.


\section{Limitations}
In this work, we investigate using iterative planning and search, mimicking the manner of our human beings, to enhance the innovation capability of existing LLM-based methods.
Despite promising findings, some limitations remain in this work, which we discuss below:

\noindent\textbf{Limited Iterations Steps.} Although our approach can significantly enhance the novelty and diversity of generated ideas through iteration, we do not see a continuous increment in generating new ideas after 3 rounds of iteration. 

\noindent\textbf{Planning without Rewards.}
In our planning and search framework, we do not introduce reward functions but only use the internal knowledge of LLMs to generate search plans. This may limit the effectiveness of planning.

We hope these findings inspire future investigations into using LLM to comprehensively integrate both
internal and external knowledge for LLM-based scientific innovation.
We believe addressing each of these shortcomings will lead to exciting future directions.

\section{Ethics Statement}

\textbf{Publication Policy}. The increasing use of AI to generate research ideas poses significant challenges to academic integrity.  The growing accessibility of LLMs and the rising usefulness of LLMs in research may lead to deterioration in the overall quality of scholarly content, as individuals may rely on AI for both creativity and submission reviews. Therefore, there is a legitimate concern that students or researchers would exploit these technologies and present low-quality research proposals. To mitigate these risks, it is crucial to hold accountability for outputs generated through AI tools in scientific submissions. 

\noindent\textbf{Intellectual Credit}. Generative AI in the research cycle poses great concerns about intellectual credit on the submitted works. While traditional frameworks were more like a tool for human researchers, LLMs are more potent in a way that plays a more significant role in the scientific research process if used. It is still unclear how intellectual credit should be distributed in the case of AI-driven research. To better attribute credit to AI-supported research, researchers should adopt transparent documentation about their research process, including the extent of AI involvement in generating ideas and developing experiments. 

\noindent\textbf{Potential for Misuse}. AI-generated research ideas, particularly those introducing novel concepts, possess the potential for misuse. This could lead to harmful outcomes. Ideation agents may be exploited to develop adversarial attack strategies or other unethical applications. Therefore, it is important to develop anti-jailbreak mechanisms or safety checks on AI-generated content and the use of generative AI in research. 

\noindent\textbf{Idea Homogenization}. If AI was widely used in scientific research, this would raise concerns about the potential idea of homogenization. The wide adoption of LLMs in research could reflect a narrower set of perspectives or systematic biases compared to human researchers not using AI assistance. Therefore, it is important to recognize the limitations of current LLM-generated ideas, and future work should focus more on enhancing the generation diversity either by improving the models themselves or by refining the ideation process.

\noindent\textbf{Impact on Human Researchers}. The challenge posed by AI's integration into research should be well recognized because research is fundamentally and historically a community-driven and collaborative effort. It is still unclear on the negative consequences of the introduction of AI in the research process. People should be cautious and aware of the potential decline in human thought and a reduction in opportunities for human collaboration after the introduction of AI in research. Future works should explore other methods of human-AI collaboration. Understanding how LLM should be integrated into the research process will be an ongoing problem. 

\bibliography{acl_latex}

\appendix

\section{Human Annotation}
This section provides details about human annotation in our human evaluation experiment.

\subsection{Annotation Instructions}
The complete annotation instruction for the idea reviewer is given below:

Please evaluate the given twenty ideas based on four criteria (Novelty, Feasibility, Effectiveness, and Overall), identify the best four and the worst four ideas, and rank them accordingly. 
The principles include:

\noindent\textbf{Annotation Dimensions:}
See Table \ref{tab:online_instructions} for detailed dimension instructions.

\begin{table*}[!htbp]
    \centering
    \begin{tabularx}{\textwidth}{lX}
         \hline
         \textbf{Criteria} & \textbf{Definition} \\
         \hline
         Novelty & Novelty refers to the originality and innovativeness of the idea. It assesses how new and unique the idea is compared to existing work in the field. \\
         Overall & Overall evaluates the general quality and potential of the idea, taking into account all other criteria (Novelty, Feasibility, and Effectiveness). It provides a holistic assessment of the idea's value. \\
         Feasibility & Feasibility assesses the practicality and implementability of the idea. It considers whether the idea can be realistically executed with available resources and within a reasonable timeframe. \\
         Effectiveness & Effectiveness evaluates the expected impact and success of the idea in achieving its intended goals. It considers how well the idea is likely to perform in practice. \\
         \hline
    \end{tabularx}
    \caption{Evaluation Criteria and Definitions for Online Idea Assessment Based on Novelty, Feasibility, Effectiveness, and Overall Quality}
    \label{tab:online_instructions}
\end{table*}

\noindent\textbf{Annotation Method:} Annotate the best 4 and the worst 4 ideas in each of the 4 dimensions. For the best ideas, mark them as 1, 2, 3, and 4 respectively, while 1 refers to the best idea. For the worst ideas, mark them as 17, 18, 19, and 20, while 20 refers to the worst idea. No need to annotate ideas other than the best 4 and the worst 4. For an example, see Table \ref{tab:online_rank_example}.
\begin{table*}[!htbp]
    \centering
    \begin{tabular}{cc}
        \hline
         Rank & Label \\
        \hline
         Best & 1 \\
         Second Best & 2 \\
         Third Best & 3 \\
         Fourth Best & 4 \\
         \ldots & \ldots \\
         \ldots & \ldots \\
         Fourth Worst & 17 \\
         Third Worst & 18 \\
         Second Worst & 19 \\
         Worst & 20 \\
        \hline
    \end{tabular}
    \caption{An example of Ranking Labels for Annotating the Best and Worst Ideas Across Four Evaluation Dimensions}
    \label{tab:online_rank_example}
\end{table*}

\subsection{Data Description}
We manually selected 20 different papers from ACL2024, CVPR2024, and ICLR2024. Each paper is carefully selected, and the 20 papers are from various research fields, including Natural Language Processing, Computer Vision, and Large Language Models in general, and have varied academic significance measured by citations to represent a broad scope of research papers. The online pilot study gives the human evaluators a form of twenty rows and five columns, along with a hyperlink to the original paper the ideas are generated. Each row is of an idea generated by one of the four different methods, Nova, AI-Scientist, AI-Researcher, or ResearchAgent. Still, human experts have no information on which method to generate that particular idea. The four columns are summary, novelty, feasibility, effectiveness, and overall. The summary is the research plan generated from one of the four methods, and the remaining four columns are entries for human experts to input their rankings. The four best ideas are labeled as 1, 2, 3, or 4, and the four worst ideas are labeled as 17, 18, 19, or 20. The rest of the entries are left blank. 

\subsection{Risk Statement}

\noindent\textbf{Physical Risk.} This study does not involve any activities that may cause physical harm or discomfort.

\noindent\textbf{Psychological Risk.} This study does not involve sensitive topics or psychological experiments.

\noindent\textbf{Social Risk.} This study does not involve activities that could affect participants' social relationships or reputations. No personal information will be disclosed.

\noindent\textbf{Economic Risk.} This study will not result in any economic loss for the participants.

\noindent\textbf{Privacy and Data Security Risk.} All annotation data will be randomly assigned to anonymous experts. No personally sensitive information will be collected.






\section{Prompts and Examples}
This section provides a comprehensive overview of various prompts and examples Nova uses in idea generation and research proposal creation. The tables are organized in Tab. \ref{table of tables} to guide the reader through different methods and stages of idea development, research trend exploration, and proposal drafting, with each table focusing on a distinct aspect of the research process.  

\begin{table*}[h]
    \centering
    \caption{Tables in the Appendix}
    \begin{tabular}{ll}
        \hline
        \textbf{Table Number} & \textbf{Table Title} \\
        \hline
        \hyperref[tab:table2]{Table 6} & Prompt for initial seed idea generation using inner knowledge from LLM \\
        \hyperref[tab:table3a]{Table 7} & An example of the initial seed idea generation using inner knowledge from LLM (Part 1) \\
        \hyperref[tab:table3b]{Table 8} & An example of the initial seed idea generation using inner knowledge from LLM (Part 2) \\
        \hyperref[tab:table4]{Table 9} & Prompt for generating  research trends\\
        \hyperref[tab:table5]{Table 10} & Current hot research trends in AI (Part 1) \\
        \hyperref[tab:table6]{Table 11} & Current hot research trends in AI (Part 2)\\
 \hyperref[tab:table8]{Table 14}&Prompt for generating research ideas based on popular research trends\\
        \hyperref[tab:table7a]{Table 12} & An example of generating research ideas based on popular research trends (Part 1)\\
        \hyperref[tab:table7b]{Table 13} & An example of generating research ideas based on popular research trends (Part 2)\\
        \hyperref[tab:table10]{Table 17} & General theory of scientific discovery (Part 1) \\
        \hyperref[tab:table11]{Table 18} & General theory of scientific discovery (Part 2) \\
        \hyperref[tab:table12a]{Table 19} & Prompt for idea generation based on the general theory of scientific discovery (Part 1) \\
        \hyperref[tab:table12b]{Table 20} & Prompt for idea generation based on the general theory of scientific discovery (Part 2) \\
        \hyperref[tab:table13a]{Table 21} & An example of idea generation based on the general theory of scientific discovery (Part 1) \\
        \hyperref[tab:table13b]{Table 22} & An example of idea generation based on the general theory of scientific discovery (Part 2) \\
        \hyperref[tab:table14]{Table 23} & Prompt for expanding idea generation using retrieved knowledge and reflection iteration \\
        \hyperref[tab:table15]{Table 24} & An example of original idea \\
        \hyperref[tab:table16]{Table 25} & An example of iterative research ideas \\
        \hyperref[tab:table17]{Table 26} & Prompt for initial proposal generation\\
        \hyperref[tab:table18]{Table 27} & An example of initial proposal \\
        \hyperref[tab:table19]{Table 28} & Prompt for method decomposition \\
        \hyperref[tab:table20]{Table 29} & An example of method decomposition (Part 1) \\
        \hyperref[tab:table21]{Table 30} & An example of method decomposition (Part 2) \\
        \hyperref[tab:table22a]{Table 31} & Prompt for final proposal generation (Part 1) \\
        \hyperref[tab:table22b]{Table 32} & Prompt for final proposal generation (Part 2) \\
        \hyperref[tab:table23]{Table 33} & An example of final proposal (Part 1) \\
        \hyperref[tab:table24]{Table 34} & An example of final proposal (Part 2) \\
        \hyperref[tab:table25]{Table 35} & Prompt for search plan generation \\
        \hyperref[tab:table26]{Table 36} & An example of search plan \\
        \hyperref[tab:table27]{Table 37} & An example of search result \\
        \hline
    \end{tabular}
    \label{table of tables}
\end{table*}

\begin{table*}[ht]
\centering
\begin{tabular}{p{12cm}}
\toprule
\textbf{Prompt} \\
\midrule
\textbf{Role:} You are an expert researcher in AI. Your goal is to propose some innovative and valuable research ideas based on the target paper. \\
\textbf{Skill:} Generate subsequent exploration ideas according to the following steps: \\
Understanding of the target paper and related papers is essential: \\
- The target paper is the primary research study you aim to enhance or build upon through future research, serving as the central source and focus for identifying and developing the specific research idea. \\
- The referenced papers are studies that the target paper has cited, indicating their direct relevance and connection to the primary research topic you are focusing on, and providing additional context and ideas that are essential for understanding and expanding upon the target paper. \\
\textbf{Step 1:} Combine target paper and referenced paper to answer the following information: \\
1. What are the tasks, methods, and main innovations of the current paper? \\
2. What are the weaknesses and limitations of the current paper? \\
\textbf{Step 2:} Propose some valuable and new research ideas. \\
\textbf{Output Format:} \{qa\_info\_with\_idea\_json\_format\} \\
\textbf{Requirements:} \\
1. Output about 5 new ideas worth exploring. \\
2. Aim for new research ideas that can potentially win best paper awards at top conferences like ACL, NeurIPS, ICLR, and CVPR. \\
3. Please output your thought process. \\
4. Please think step by step. \\
\textbf{Input:} \\
target\_title: \{paper.title\} \\
target\_abstract: \{paper.abstract\} \\
referenced\_paper\_title: \{paper.related\_paper\_titles\} \\
referenced\_paper\_abstract: \{paper.related\_paper\_abstract\} \\
\textbf{Output:} \\
Thinking: \\
\textless output your thinking process here, explain why it should have changed to win the best paper awards at top conferences\textgreater \\
New research idea that improves target\_paper: \\
\textless JSON\textgreater \\
\midrule
\bottomrule
\end{tabular}
\caption{Prompt for initial seed idea generation using inner knowledge from LLM.}
\label{tab:table2}
\end{table*}

\begin{table*}[ht]
\centering
\begin{tabular}{p{12cm}}
\toprule
\textbf{ResearchAgent: Iterative Research Idea Generation over Scientific Literature with Large Language Models} \\ \midrule
\textbf{Generated Research Ideas (Part 1)} \\ \midrule

1. \textbf{Integrate Real-Time Data Feeds} \\
   \textit{Thinking:} The target paper's reliance on existing literature may limit the generation of truly novel ideas. By incorporating real-time data sources, such as ongoing research developments or recent publications, the ResearchAgent can generate more innovative and timely research ideas. \\
   \textit{Idea:} Integrate real-time data feeds and recent publications into the ResearchAgent’s knowledge base to ensure that the generated ideas are aligned with the latest scientific advancements. \\
   \textit{Keywords:} real-time data integration, recent publications tracking, innovative research ideas \\ \midrule

2. \textbf{Develop a Multi-Faceted Evaluation Framework} \\
   \textit{Thinking:} The evaluation metrics used in the target paper may not fully capture the practical applicability and impact of the generated ideas. By incorporating domain-specific expert reviews and practical implementation feedback, the evaluation process can be more comprehensive. \\
   \textit{Idea:} Develop a multi-faceted evaluation framework that includes domain-specific expert reviews and practical implementation feedback to assess the applicability and impact of the generated research ideas. \\
   \textit{Keywords:} multi-faceted evaluation, expert reviews, practical implementation feedback \\ \midrule

3. \textbf{Optimize Computational Efficiency} \\
   \textit{Thinking:} The iterative refinement process in the target paper might be computationally expensive and time-consuming. By optimizing the computational efficiency of the ReviewingAgents and leveraging parallel processing techniques, the scalability of the ResearchAgent can be improved. \\
   \textit{Idea:} Optimize the computational efficiency of the ReviewingAgents and leverage parallel processing techniques to enhance the scalability of the ResearchAgent. \\
   \textit{Keywords:} computational efficiency, parallel processing, scalability \\ \midrule

\bottomrule
\end{tabular}
\caption{An example of seed idea generation using inner knowledge from LLM (Part 1).}
\label{tab:table3a}
\end{table*}

\begin{table*}[ht]
\centering
\begin{tabular}{p{12cm}}
\toprule
\textbf{ResearchAgent: Iterative Research Idea Generation over Scientific Literature with Large Language Models} \\ \midrule
\textbf{Generated Research Ideas (Part 2)} \\ \midrule

4. \textbf{Incorporate Multi-Modal Data Sources} \\
   \textit{Thinking:} The target paper focuses on generating research ideas based on scientific literature. By incorporating multi-modal data sources, such as experimental data, patents, and industry reports, the ResearchAgent can generate more comprehensive and interdisciplinary research ideas. \\
   \textit{Idea:} Incorporate multi-modal data sources, including experimental data, patents, and industry reports, into the ResearchAgent’s knowledge base to generate more comprehensive and interdisciplinary research ideas. \\
   \textit{Keywords:} multi-modal data sources, experimental data, interdisciplinary research ideas \\ \midrule

5. \textbf{Incorporate Diverse ReviewingAgents} \\
   \textit{Thinking:} The target paper uses ReviewingAgents instantiated with human preference-aligned LLMs for iterative feedback. By incorporating a diverse set of ReviewingAgents with different expertise and perspectives, the feedback process can be more robust and holistic. \\
   \textit{Idea:} Incorporate a diverse set of ReviewingAgents with different expertise and perspectives to provide more robust and holistic feedback during the iterative refinement process. \\
   \textit{Keywords:} diverse ReviewingAgents, expertise diversity, holistic feedback \\ \midrule

\bottomrule
\end{tabular}
\caption{An example of seed idea generation using inner knowledge from LLM (Part 2).}
\label{tab:table3b}
\end{table*}

\begin{table*}[ht]
\centering
\begin{tabular}{p{12cm}}
\toprule
\textbf{Prompt} \\
\midrule 
\textbf{Role:} You are an AI expert researcher. You can summarise the current hot research trends from the list of recent AI papers. \\
\textbf{Skill:} You will analyze the research trending based on the recent popular paper, provide us with the research trending report. \\
\textbf{Requirements:} 
\begin{enumerate}
    \item Provide a comprehensive analysis, including the hot research directions, the highlights of the technologies and methods, and discuss whether these technologies can be used in other fields.
\end{enumerate}
I will provide a list of recent popular paper list here: \{popular\_paper\_list\} \\
Then, Please output the current research trending report here: \\
\midrule
\bottomrule
\end{tabular}
\caption{Prompt for generating  popular research trends.}
\label{tab:table4}
\end{table*}

\begin{table*}[ht]
\centering
\begin{tabular}{p{12cm}}
\toprule
\textbf{Hot Research Directions} \\
\midrule
1. Long-Context Language Models (LLMs) and Retrieval-Augmented Generation (RAG) \\
   - \textbf{Key Papers}: "RAG in the Era of Long-Context LLMs", "LongCite", "MemLong", "Improved RAG with Self-Reasoning", "LongWriter", "EfficientRAG", "Enhanced RAG with Long-Context LLMs", "GraphReader" \\
   - \textbf{Highlights}: Addressing the challenge of maintaining focus and relevance in long-context LLMs. Combining RAG mechanisms with long-context capabilities to improve performance in tasks like question answering and citation generation. Innovations such as order-preserving RAG, external retrievers, and graph-based systems to enhance context handling. \\
   - \textbf{Cross-Field Applications}: These advancements can be applied in fields requiring extensive document analysis, such as legal research, academic literature review, and medical records analysis. \\
\midrule
2. Strategic Chain-of-Thought (CoT) and Self-Improvement Techniques \\
   - \textbf{Key Papers}: "Strategic Chain-of-Thought", "Teaching LLM Agents to Self-Improve", "Self-Taught Evaluators", "Meta-Rewarding LLMs", "SelfGoal" \\
   - \textbf{Highlights}: Incorporating strategic knowledge to guide intermediate reasoning steps. Iterative self-improvement and self-evaluation to enhance model performance over multiple turns. Use of self-generated training data to refine judgment and reasoning capabilities. \\
   - \textbf{Cross-Field Applications}: These methods can be beneficial in educational technologies, autonomous decision-making systems, and any domain requiring iterative problem-solving and learning. \\
\midrule
3. Mixture-of-Experts (MoE) and Multi-Agent Systems \\
   - \textbf{Key Papers}: "OLMoE", "Agentic RAG for Time Series Analysis", "Mixture-of-Agents", "MindSearch" \\
   - \textbf{Highlights}: Leveraging sparse Mixture-of-Experts to optimize model performance and efficiency. Multi-agent architectures for specialized task handling, such as time series analysis and complex web-information seeking. \\
   - \textbf{Cross-Field Applications}: These approaches can be utilized in financial forecasting, climate modeling, and complex system simulations where specialized expertise is crucial. \\
\midrule
4. Synthetic Data Generation and Utilization \\
   - \textbf{Key Papers}: "Smaller, Weaker, Yet Better", "Scaling Synthetic Data Creation", "Improving Retrieval in LLMs through Synthetic Data", "Model Collapse on Synthetic Data" \\
   - \textbf{Highlights}: Using weaker models to generate high-quality synthetic data for fine-tuning stronger models. Addressing the challenges of model collapse due to recursive training on synthetic data. \\
   - \textbf{Cross-Field Applications}: Synthetic data can be used in privacy-preserving data analysis, training AI models in healthcare, and augmenting datasets in low-resource languages. \\
\midrule
\bottomrule
\end{tabular}
\caption{Current hot research trends in AI (Part 1)}
\label{tab:table5}
\end{table*}

\begin{table*}[ht]
\centering
\begin{tabular}{p{12cm}}
\toprule
\textbf{Hot Research Directions} \\
\midrule
5. Controllable and Robust Text Generation \\
   - \textbf{Key Papers}: "Controllable Text Generation for LLMs", "Enhancing Robustness in LLMs", "Improving Legibility of LLM Outputs" \\
   - \textbf{Highlights}: Techniques for controlling the style, safety, and consistency of generated text. Methods to enhance robustness by filtering out irrelevant information and improving the clarity of outputs. \\
   - \textbf{Cross-Field Applications}: These advancements are crucial for developing reliable AI assistants, automated content generation, and ensuring the safety of AI-generated outputs in sensitive applications. \\
\midrule
6. AI in Scientific Discovery and Evaluation \\
   - \textbf{Key Papers}: "The AI Scientist", "Automate Design of Agentic Systems", "Self-Taught Evaluators" \\
   - \textbf{Highlights}: AI agents capable of conducting independent research and writing scientific papers. Meta-agent frameworks for designing and evaluating agentic systems. \\
   - \textbf{Cross-Field Applications}: These innovations can revolutionize scientific research, enabling faster discovery and validation of new theories across various scientific disciplines. \\
\midrule
7. Advanced Prompt Engineering and Personalization \\
   - \textbf{Key Papers}: "Conversational Prompt Engineering", "A Survey of Prompt Engineering Methods in LLMs" \\
   - \textbf{Highlights}: Techniques for creating personalized prompts through iterative user interaction. Comprehensive surveys on prompt engineering methods for various NLP tasks. \\
   - \textbf{Cross-Field Applications}: Personalized prompt engineering can enhance user experience in customer service bots, personalized education platforms, and adaptive learning systems. \\
\midrule
8. AI in Code and Software Engineering \\
   - \textbf{Key Papers}: "LLM Compiler", "From LLMs to LLM-based Agents for Software Engineering", "DeepSeek-Coder-V2" \\
   - \textbf{Highlights}: Models designed for code optimization and generation. Surveys on the application of LLMs in software engineering tasks like requirement engineering and test generation. \\
   - \textbf{Cross-Field Applications}: These advancements can improve software development workflows, automate code reviews, and enhance the capabilities of integrated development environments (IDEs). \\
\bottomrule
\end{tabular}
\caption{Current hot research trends in AI (Part 2)}
\label{tab:table6}
\end{table*}

\begin{table*}[ht]
\centering
\begin{tabular}{p{12cm}}
\toprule
\textbf{ResearchAgent: Iterative Research Idea Generation over Scientific Literature with Large Language Models} \\ \midrule
\textbf{Generated Research Ideas from Popular Knowledge (Part 1)} \\ \midrule

1. \textbf{Integrate Synthetic Data Generation Techniques} \\
   \textit{Thinking:} The target paper relies heavily on existing literature, which may limit the generation of truly novel ideas. By integrating synthetic data generation techniques, we can create high-quality synthetic research papers and datasets to augment the existing literature. This approach can help overcome the dependency on existing literature and introduce more diversity in the generated ideas. \\
   \textit{Idea:} Integrate synthetic data generation techniques to create high-quality synthetic research papers and datasets, augmenting the existing literature and introducing more diversity in the generated ideas. \\
   \textit{Keywords:} synthetic data generation, research idea generation, LLMs, data augmentation \\
   \textit{Rationale:} This idea addresses the limitation of dependency on existing literature by introducing synthetic data, which can diversify the input data and potentially lead to more novel research ideas. The use of synthetic data generation is innovative and has the potential to enhance the robustness and creativity of the ResearchAgent, making it a strong candidate for top conference recognition. \\ \midrule

2. \textbf{Incorporate Strategic Chain-of-Thought and Self-Improvement Techniques} \\
   \textit{Thinking:} The evaluation metrics used in the target paper may not fully capture the practical applicability and impact of the generated ideas. By incorporating strategic chain-of-thought and self-improvement techniques, we can enhance the evaluation process by enabling the ReviewingAgents to iteratively refine their feedback and improve their judgment over multiple turns. \\
   \textit{Idea:} Incorporate strategic chain-of-thought and self-improvement techniques to enhance the evaluation process, enabling ReviewingAgents to iteratively refine their feedback and improve their judgment over multiple turns. \\
   \textit{Keywords:} strategic chain-of-thought, self-improvement, iterative feedback, evaluation metrics \\
   \textit{Rationale:} This idea improves the evaluation process by making it more dynamic and iterative, allowing ReviewingAgents to learn and refine their feedback over time. This approach can lead to more accurate and practical evaluations of the generated research ideas, addressing a key limitation of the target paper. The innovative use of self-improvement techniques makes this idea a strong contender for top conference recognition. \\

\bottomrule
\end{tabular}
\caption{An example of generating research ideas based on popular research trends (Part 1).}
\label{tab:table7a}
\end{table*}

\begin{table*}[ht]
\centering
\begin{tabular}{p{12cm}}
\toprule
\textbf{ResearchAgent: Iterative Research Idea Generation over Scientific Literature with Large Language Models} \\ \midrule
\textbf{Generated Research Ideas from Popular Knowledge (Part 2)} \\ \midrule

3. \textbf{Leverage Mixture-of-Experts and Multi-Agent Systems} \\
   \textit{Thinking:} The iterative refinement process in the target paper might be computationally expensive and time-consuming. By leveraging mixture-of-experts (MoE) and multi-agent systems, we can optimize the performance and efficiency of the ResearchAgent, making the iterative refinement process more scalable. \\
   \textit{Idea:} Leverage mixture-of-experts (MoE) and multi-agent systems to optimize the performance and efficiency of the ResearchAgent, making the iterative refinement process more scalable. \\
   \textit{Keywords:} mixture-of-experts, multi-agent systems, scalability, performance optimization \\
   \textit{Rationale:} This idea addresses the scalability limitation by optimizing the computational efficiency of the ResearchAgent using MoE and multi-agent systems. This approach can significantly reduce the time and resources required for the iterative refinement process, making the system more scalable and practical for large-scale applications. The innovative use of MoE and multi-agent systems enhances the feasibility and impact of the ResearchAgent, making it a strong candidate for top conference recognition. \\

\bottomrule
\end{tabular}
\caption{An example of generating research ideas based on popular research trends (Part 2).}
\label{tab:table7b}
\end{table*}

\begin{table*}[ht]
\centering
\begin{tabular}{p{12cm}}
\toprule
\textbf{Prompt} \\
\midrule
\textbf{Role:} You are an expert researcher in AI. Your goal is to propose some innovative and valuable research ideas based on the target paper and some high-quality research trends. \\

\textbf{Skills:} Propose some innovative and valuable research ideas following these steps: \\
1. Understand the target paper and target\_paper\_base\_info well. \\
2. Understand the research\_trending\_info and high\_quality\_paper\_list well. Analyze the latest innovations, ideas, and methods they have used. \\
3. List some technologies used in research\_trending\_info and high\_quality\_paper\_list, analyze the feasibility of combining these technologies with target\_paper\_base\_info, and analyze the advantages and disadvantages of the combination. \\
4. Come up with some innovative and valuable research ideas. \\

\textbf{Input Data Description:} It is important to understand target\_paper\_base\_info and research\_trending\_info: \\
- target\_paper\_base\_info provides some basic information and limitation information of the target paper. \\
- research\_trending\_info and high\_quality\_paper\_list: This data analyzes the current popular research trends and high-quality papers. You can get inspiration from them and draw on their latest methods, ideas, and innovations to come up with new ideas, but you must be different from them and make sure that the ideas you propose are reasonable. \\
- exist\_idea is some ideas that have been proposed before; you need to propose some different ones. \\

\textbf{Requirements:} \\
1. Output about 5 new ideas worth exploring. \\
2. You should aim for new research ideas that can potentially win best paper awards at top conferences like ACL, NeurIPS, ICLR, and CVPR. \\
3. The research trending may not well match the target paper; the ideas you make should make sense and be reasonable. Explain which trending or high-quality paper your inspiration comes from. What is the original idea, what are the benefits of adopting this idea, and explain why it is feasible. \\
4. Please output your thought process. \\
5. Please think step by step. \\

\textbf{Input:} \\
target paper title: \{paper.title\} \\
target paper abstract: \{paper.abstract\} \\
target\_paper\_base\_info: \{target\_paper\_base\_info\} \\
research\_trending\_info: \{research\_trending\_info\} \\
high\_quality\_paper\_list: \{str(topk\_high\_quality\_paper\_list)\} \\
exist\_idea: \{exist\_idea\} \\
target\_paper\_base\_info: \{target\_paper\_base\_info\} \\

\textbf{Output:} \\
Thinking: \\
\textless output your thinking process here, explaining what new knowledge you used for the new idea and why it makes sense and why it should have changed to win the best paper awards at top conferences\textgreater \\
New research idea that improves target\_paper: \\
\textless JSON\textgreater \\
\midrule
\bottomrule
\end{tabular}
\caption{Prompt for generating research ideas Based on recent high-quality articles.}
\label{tab:table8}
\end{table*}

\begin{table*}[ht]
\centering
\begin{tabular}{p{12cm}}
\toprule
\textbf{1. Idea: Integrate Retrieval-Augmented Generation (RAG)} \\
\textit{Thinking:} To address the limitation of handling abstract or ambiguous user prompts, we can leverage the concept of Retrieval-Augmented Generation (RAG) from the latest research trends. By incorporating an external retriever, we can enhance the model's ability to understand and generate more accurate visual representations. \\
\textit{Keywords:} Retrieval-Augmented Generation, RAG, text-to-image generation, abstract prompts. \\
\textit{Rationale:} RAG can help retrieve relevant information from external sources, thereby improving the model's ability to handle abstract or ambiguous prompts. This approach is innovative as it combines retrieval mechanisms with text-to-image generation, potentially leading to more accurate and expressive visual outputs. \\
\textit{Source:} popular\_knowledge \\

\textbf{2. Idea: Implement Sparse Mixture-of-Experts (MoE) Models} \\
\textit{Thinking:} To improve scalability and handle a growing number of users, we can utilize Sparse Mixture-of-Experts (MoE) models. These models can dynamically allocate computational resources, making the system more efficient and scalable. \\
\textit{Keywords:} Sparse Mixture-of-Experts, MoE, scalability, text-to-image generation. \\
\textit{Rationale:} MoE models are known for their efficiency and scalability, as they activate only a subset of experts for each input. This can help manage the computational load and improve the system's performance as the number of users grows. \\
\textit{Source:} popular\_knowledge \\

\textbf{3. Idea: Incorporate Strategic and Chain-of-Thought Reasoning} \\
\textit{Thinking:} To enhance the expressiveness and alignment of user prompts, we can incorporate Strategic and Chain-of-Thought Reasoning. This approach can help break down complex prompts into intermediate steps, leading to better understanding and generation. \\
\textit{Keywords:} Chain-of-Thought Reasoning, strategic reasoning, text-to-image generation, complex prompts. \\
\textit{Rationale:} Chain-of-Thought Reasoning can help decompose complex prompts into manageable steps, improving the model's ability to generate accurate and expressive visual outputs. This approach is innovative as it introduces a new way of handling complex prompts, potentially leading to significant improvements in performance. \\
\textit{Source:} popular\_knowledge \\
\bottomrule
\end{tabular}
\caption{An example of generating research ideas based on recent high-quality articles (Part 1).}
\label{tab:table9a}
\end{table*}

\begin{table*}[ht]
\centering
\begin{tabular}{p{12cm}}
\toprule
\textbf{4. Idea: Explore Meta Learning Techniques} \\
\textit{Thinking:} To address the limitation of generalizing to new or unseen user preferences, we can explore the use of Meta Learning. This approach can help the model quickly adapt to new user preferences based on limited interaction data. \\
\textit{Keywords:} Meta Learning, adaptation, user preferences, text-to-image generation. \\
\textit{Rationale:} Meta Learning can help the model generalize better to new or unseen user preferences by learning how to learn from limited data. This approach is innovative as it introduces a new way of improving the model's adaptability, potentially leading to more personalized and accurate visual outputs. \\
\textit{Source:} popular\_knowledge \\

\textbf{5. Idea: Adopt Techniques from Controllable Text Generation} \\
\textit{Thinking:} To improve the robustness and controllability of text generation, we can adopt techniques from Controllable Text Generation. This can help ensure that the generated visual outputs are consistent with user preferences and safe for various applications. \\
\textit{Keywords:} Controllable Text Generation, robustness, consistency, text-to-image generation. \\
\textit{Rationale:} Controllable Text Generation techniques can help manage the style, safety, and consistency of generated outputs. This approach is innovative as it introduces a new layer of control over the generation process, potentially leading to more reliable and user-aligned visual outputs. \\
\textit{Source:} popular\_knowledge \\
\bottomrule
\end{tabular}
\caption{An example of generating research ideas based on recent high-quality articles (Part 2).}
\label{tab:table9b}
\end{table*}

\begin{table*}[ht]
\label{sci-discovery-theory part-I}
\centering
\begin{tabular}{p{14cm}}
\toprule
\textbf{General theory of scientific discovery 1-5} \\ \midrule

\textbf{1. Define New Scientific Problems} \\
\textbf{Theoretical Basis:} Kuhn's paradigm theory, Laudan's problem-solving model, Nichols's problem-generation theory. \\
\textbf{Method:} Identify anomalies in existing theories; explore theoretical boundaries and scope of application; integrate interdisciplinary knowledge and discover new problems; re-examine neglected historical problems. \\ \midrule

\textbf{2. Propose New Hypotheses} \\
\textbf{Theoretical Basis:} Pierce's hypothetical deduction method, Weber's theory of accidental discovery, Simon's scientific discovery as problem solving. \\
\textbf{Method:} Analogical reasoning; thought experiment; intuition and creative leaps; reductio ad absurdum thinking. \\ \midrule

\textbf{3. Exploring the Limitations and Shortcomings of Current Methods} \\
\textbf{Theoretical Basis:} Popper's falsificationism, Lakatos's research program methodology, Feyerabend's methodological anarchism. \\
\textbf{Method:} Critically analyze existing methods; find deviations between theoretical predictions and experimental results; explore the performance of methods under extreme conditions; interdisciplinary comparative methodology. \\ \midrule

\textbf{4. Design and Improve Existing Methods} \\
\textbf{Theoretical Basis:} Laudan's methodological improvement model, Ziemann's creative extension theory, Hacking's experimental system theory. \\
\textbf{Method:} Integrate new technologies and tools; improve experimental design and control; improve measurement accuracy and resolution; develop new data analysis methods. \\ \midrule

\textbf{5. Abstract and Summarize the General Laws Behind Multiple Related Studies} \\
\textbf{Theoretical Basis:} Whewell's conceptual synthesis theory, Carnap's inductive logic, Glaser and Strauss's grounded theory. \\
\textbf{Method:} Comparative analysis of multiple case studies; identify common patterns and structures; construct conceptual frameworks and theoretical models; formal and mathematical descriptions. \\ \midrule

\bottomrule
\end{tabular}
\caption{General theory of scientific discovery (Part 1)}
\label{tab:discovery_method_1}
\end{table*}

\begin{table*}[ht]
\label{sci-discovery-theory part-II}
\centering
\begin{tabular}{p{14cm}}
\toprule
\textbf{General theory of scientific discovery 6-10} \\ \midrule

\textbf{6. Construct and Modify Theoretical Models} \\
\textbf{Theoretical Basis:} Quine's holism, Lakoff's conceptual metaphor theory, Kitcher's unified theory of science. \\
\textbf{Method:} Form a balance between reductionism and emergence; develop an interdisciplinary theoretical framework; mathematical modeling and computer simulation; theoretical simplification and unification. \\ \midrule

\textbf{7. Designing Critical Experiments} \\
\textbf{Theoretical Basis:} Duhem-Quine thesis, Bayesian experimental design theory, Mayo's experimental reasoning theory. \\
\textbf{Method:} Designing experiments that can distinguish competing theories; exploring extreme conditions and boundary cases; developing new observation and measurement techniques; designing natural experiments and quasi-experiments. \\ \midrule

\textbf{8. Explaining and Integrating Anomalous Findings} \\
\textbf{Theoretical Basis:} Hansen's theory of anomalous findings, Sutton's model of scientific serendipity, Kuhn's theory of crises and revolutions. \\
\textbf{Method:} Revisiting basic assumptions; developing auxiliary hypotheses; exploring new explanatory frameworks; integrating multidisciplinary perspectives. \\ \midrule

\textbf{9. Evaluating and Selecting Competing Theories} \\
\textbf{Theoretical Basis:} Reichenbach's confirmation theory, Sober's theory selection criteria, Laudan's problem-solving progress assessment. \\
\textbf{Method:} Comparing theories for explanatory power and predictive power; evaluating the simplicity and elegance of theories; considering the heuristics and research agenda of theories; weighing the empirical adequacy and conceptual coherence of theories. \\ \midrule

\textbf{10. Scientific Paradigm Shift} \\
\textbf{Theoretical Basis:} Kuhn's theory of scientific revolutions, Toulmin's model of conceptual evolution, Hall's dynamic system theory. \\
\textbf{Method:} Identify accumulated anomalies and crises; develop new conceptual frameworks; reinterpret and organize known facts; establish new research traditions and practices. \\ \midrule

\bottomrule
\end{tabular}
\caption{General theory of scientific discovery (Part 2)}
\label{tab:discovery_method_2}
\end{table*}

\begin{table*}[ht]
\centering
\begin{tabular}{p{12cm}}
\toprule
\textbf{Prompt} \\
\midrule
\textbf{Role:} You are an expert researcher in AI. You are familiar with Science Discovery Theory, and you can use these theories to propose some innovative and valuable research ideas based on the information provided by users. \\

\textbf{Skill:} Follow the steps below to generate new ideas and ideas for exploration: \\
1. Understanding of the target paper and related papers is essential:
    \begin{itemize}
        \item The target paper is the primary research study you aim to enhance or build upon through future research, serving as the central source and focus for identifying and developing the specific research idea.
        \item The referenced papers are studies that the target paper has cited, indicating their direct relevance and connection to the primary research topic you are focusing on, and providing additional context and ideas that are essential for understanding and expanding upon the target paper.
    \end{itemize}
2. Understanding of the science discovery theories is essential: You need to select appropriate theories and combine the information provided by the current paper to come up with creative, influential, and feasible ideas. \\
3. Here are 10 general laws and methodologies of scientific discovery from the perspective of the philosophy of science. You can choose one or more of these methodologies and propose new scientific research ideas for the target paper:
    \{self.scientific\_discovery\_theory\} \\
\bottomrule
\end{tabular}
\caption{Prompt for idea generation based on general theory of scientific discovery (Part 1).}
\label{tab:table12a}
\end{table*}

\begin{table*}[ht]
\centering
\begin{tabular}{p{12cm}}
\toprule
4. Select 5 most appropriate theories and methods that are most suitable for the target paper and put forward 5 new ideas. \\

\textbf{Requirements:}
\begin{enumerate}
    \item Output about 5 new ideas worth exploring.
    \item You should aim for new research ideas that can potentially win best paper awards at top conferences like ACL, NeurIPS, ICLR, and CVPR.
    \item Skip the research theories that may not well match the target paper; the theory and method you use should make sense and be reasonable for the target paper.
    \item Thinking is your thinking process. Please explain which theory you used in your thinking process.
    \item Please output your thought process.
    \item Please think step by step.
\end{enumerate}

\textbf{Input:}
\begin{itemize}
    \item Referenced papers titles: \{paper.related\_paper\_titles\}
    \item Referenced papers abstracts: \{paper.related\_paper\_abstract\}
    \item Target paper title: \{paper.title\}
    \item Target paper abstract: \{paper.abstract\}
\end{itemize}

\textbf{Output:} \\
\textit{Thinking:} \\
\textless output your thinking process here, explain why you chose these theories to discover new ideas and why it should have a chance to win the best paper awards at top conferences\textgreater \\
\midrule
\bottomrule
\end{tabular}
\caption{Prompt for idea generation based on general theory of scientific discovery (Part 2).}
\label{tab:table12b}
\end{table*}

\begin{table*}[ht]
\centering
\begin{tabular}{p{12cm}}
\toprule
\textbf{ResearchAgent: Iterative Research Idea Generation over Scientific Literature with Large Language Models} \\ \midrule
\textbf{Generated Research Ideas Using Science Discovery Theories (Part 1)} \\ \midrule

1. \textbf{Integrate real-time data feeds and recent publications into the ResearchAgent’s knowledge base to ensure that the generated ideas are aligned with the latest scientific advancements.} \\
   \textit{Thinking:} Using Kuhn's paradigm theory and Laudan's problem-solving model, I identified the anomaly that the current ResearchAgent does not fully utilize real-time data feeds and recent publications. By integrating real-time data feeds and recent publications, we can ensure that the generated ideas are aligned with the latest scientific advancements. This idea is innovative because it addresses the gap in the current model's ability to stay updated with the latest research, and it is feasible because real-time data integration is a well-established technique. \\
   \textit{Idea:} Integrate real-time data feeds and recent publications into the ResearchAgent’s knowledge base to ensure that the generated ideas are aligned with the latest scientific advancements. \\
   \textit{Keywords:} real-time data integration, recent publications, ResearchAgent, scientific advancements \\
   \textit{Rationale:} This idea solves the challenge of keeping the ResearchAgent updated with the latest research, making the generated ideas more relevant and impactful. It is innovative because it leverages real-time data integration, and it has the potential to win best paper awards due to its practical applicability and novelty. \\ \midrule

2. \textbf{Develop a multi-faceted evaluation framework that includes domain-specific expert reviews and practical implementation feedback to assess the applicability and impact of the generated research ideas.} \\
   \textit{Thinking:} Using Pierce's hypothetical deduction method and Weber's theory of accidental discovery, I propose a new hypothesis that the ResearchAgent can be enhanced by incorporating domain-specific expert reviews and practical implementation feedback. This idea is innovative because it introduces a multi-faceted evaluation framework that includes both expert reviews and practical feedback, making the generated research ideas more robust and applicable. \\
   \textit{Idea:} Develop a multi-faceted evaluation framework that includes domain-specific expert reviews and practical implementation feedback to assess the applicability and impact of the generated research ideas. \\
   \textit{Keywords:} multi-faceted evaluation framework, domain-specific expert reviews, practical implementation feedback, ResearchAgent \\
   \textit{Rationale:} This idea addresses the challenge of evaluating the applicability and impact of the generated research ideas. It is innovative because it combines expert reviews with practical feedback, and it has the potential to win best paper awards due to its comprehensive approach to evaluation. \\

\bottomrule
\end{tabular}
\caption{An example of idea generation based on general theory of scientific discovery (Part 1).}
\label{tab:table13a}
\end{table*}

\begin{table*}[ht]
\centering
\begin{tabular}{p{12cm}}
\toprule
\textbf{ResearchAgent: Iterative Research Idea Generation over Scientific Literature with Large Language Models} \\ \midrule
\textbf{Generated Research Ideas Using Science Discovery Theories (Part 2)} \\ \midrule

3. \textbf{Optimize the computational efficiency of the ReviewingAgents and leverage parallel processing techniques to enhance the scalability of the ResearchAgent.} \\
   \textit{Thinking:} Using Popper's falsificationism and Lakatos's research program methodology, I critically analyzed the existing methods and identified the limitation that the current ReviewingAgents are not optimized for computational efficiency. By optimizing the computational efficiency of the ReviewingAgents and leveraging parallel processing techniques, we can enhance the scalability of the ResearchAgent. \\
   \textit{Idea:} Optimize the computational efficiency of the ReviewingAgents and leverage parallel processing techniques to enhance the scalability of the ResearchAgent. \\
   \textit{Keywords:} computational efficiency, ReviewingAgents, parallel processing, scalability \\
   \textit{Rationale:} This idea solves the challenge of scalability in the ResearchAgent model. It is innovative because it focuses on optimizing computational efficiency, and it has the potential to win best paper awards due to its practical implications for large-scale research idea generation. \\

\bottomrule
\end{tabular}
\caption{An example of idea generation based on general theory of scientific discovery (Part 2).}
\label{tab:table13b}
\end{table*}

\begin{table*}[ht]
\centering
\begin{tabular}{p{12cm}}
\toprule
\textbf{Prompt} \\
\midrule
\textbf{Role:} You are an expert researcher in AI. You can learn from new knowledge and provide some impactful and creative new ideas based on user input. \\

\textbf{Skills:} Propose some innovative and valuable new research ideas following these steps: \\
1. Understand the target paper and old idea well. \\
2. Understand the new knowledge well, analyze the innovations, ideas, and methods they have used, thinking about what can be used to propose new ideas. \\
3. Generate 3 most innovative and important ideas at the end. \\

\textbf{Input Data Description:} It is important to understand target paper, old idea, and new knowledge: \\
- The target paper is the primary research study you aim to enhance or build upon through future research, serving as the central source and focus for identifying and developing the specific research idea. \\
- The old idea is an existing idea. Your new idea should be different from the old. \\
- The new knowledge consists of some literature and papers you read when you studied the old idea. Your new ideas should be inspired by the new knowledge. \\

\textbf{Requirements:} \\
1. The new idea should be related to the target paper but different from the old idea. \\
2. You should aim for new research ideas that can potentially win best paper awards at top conferences like ACL, NeurIPS, ICLR, and CVPR. \\
3. The above papers are only for inspiration, and you should not cite them; just make some incremental modifications. Instead, you should ensure your ideas are novel and distinct from the prior literature. \\
4. Please output your thought process. \\
5. Only the final three ideas should be in JSON format. \\
6. Please think step by step. \\

\textbf{Input:} \\
target paper title: \{paper.title\} \\
target paper abstract: \{paper.abstract\} \\
old\_idea: \{old\_idea\} \\
new knowledge: \{new\_knowledge\} \\

\textbf{Output:} \\
\textbf{Thinking:} \\
\textless output your thinking process here, explaining what new knowledge you used for the new idea and why it makes sense and why it should have changed to win the best paper awards at top conferences\textgreater \\
\textbf{Final Three Most Innovative and Important Ideas:} \\
\textless JSON\textgreater \\
\midrule
\bottomrule
\end{tabular}
\caption{Prompt for expanding idea generation using retrieved knowledge and reflection iteration.}
\label{tab:table14}
\end{table*}

\begin{table*}[ht]
\centering
\begin{tabular}{p{12cm}}
\toprule
\textbf{ResearchAgent: Iterative Research Idea Generation over Scientific Literature with Large Language Models} \\ \midrule
\textbf{Thinking:} The evaluation metrics used in the target paper may not fully capture the practical applicability and impact of the generated ideas. By incorporating domain-specific expert reviews and practical implementation feedback, the evaluation process can be more comprehensive. \\
\textbf{Idea:} Develop a multi-faceted evaluation framework that includes domain-specific expert reviews and practical implementation feedback to assess the applicability and impact of the generated research ideas. \\
\textbf{Keywords:} multi-faceted evaluation, expert reviews, practical implementation feedback \\
\textbf{Source:} llm\_inter\_knowledge \\
\bottomrule
\end{tabular}
\caption{An example of original idea.}
\label{tab:table15}
\end{table*}

\begin{table*}[ht]
\centering
\begin{tabular}{p{12cm}}
\toprule
\textbf{ResearchAgent: Iterative Research Idea Generation over Scientific Literature with Large Language Models} \\ \midrule
\textbf{Iterative Ethical Research Ideas} \\ \midrule

1. \textbf{Ethical Research Agent} \\
   \textit{Thinking:} Ensuring that generated research ideas adhere to ethical guidelines promotes responsible research practices and addresses the growing concern about ethical issues in AI and research. \\
   \textit{Idea:} Ethical Research Agent \\
   \textit{Keywords:} ethical AI, responsible research, ethical guidelines, research ethics \\
   \textit{Rationale:} This idea is innovative because it integrates ethical considerations into the research idea generation process, addressing a critical need for responsible AI and research practices. It has the potential to win top conference awards due to its focus on ethics and its potential to promote trust and transparency in AI-generated research ideas. \\ \midrule

2. \textbf{Ethical Value Alignment Framework for Large Language Models} \\
   \textit{Thinking:} This idea leverages insights from the new knowledge on ethical value alignment in big models. By developing a comprehensive framework, we can address the inherent ethical challenges in LLMs, ensuring that they align with societal values and norms. \\
   \textit{Idea:} Ethical Value Alignment Framework for Large Language Models \\
   \textit{Keywords:} ethical value alignment, large language models, AI ethics, normative guidelines, interdisciplinary collaboration \\
   \textit{Rationale:} This idea addresses the critical challenge of aligning ethical values in LLMs, which is essential for their responsible deployment. It is innovative because it proposes a unified framework that incorporates normative guidelines and interdisciplinary collaboration, ensuring that LLMs adhere to ethical standards. This idea has the potential to win best paper awards due to its relevance and impact on the ethical development of AI. \\ \midrule

3. \textbf{Real-time Ethical Feedback Mechanism} \\
   \textit{Thinking:} Developing a real-time ethical feedback mechanism within the ResearchAgent framework can provide immediate feedback on ethical compliance, helping researchers make informed decisions and improve the ethical alignment of their work. \\
   \textit{Idea:} Real-time Ethical Feedback Mechanism \\
   \textit{Keywords:} real-time feedback, ethical AI, immediate feedback, ethical compliance, decision-making \\
   \textit{Rationale:} This idea is innovative as it offers a practical solution to ethical challenges by providing immediate feedback, allowing researchers to address ethical issues promptly. Its potential to improve ethical compliance and decision-making makes it a strong contender for best paper awards. \\ \midrule

\bottomrule
\end{tabular}
\caption{An example of iterative research ideas.}
\label{tab:table16}
\end{table*}

\begin{table*}[ht]
\centering
\begin{tabular}{p{12cm}}
\toprule
\textbf{Prompt} \\
\midrule
\textbf You are an expert researcher in Large Language Models. Now I want you to help me brainstorm the detailed research project proposal based on the idea: \texttt{idea}. \\
The given idea is derived from paper: \texttt{paper.title}, abstract: \texttt{paper.abstract}, this is just for your background knowledge: \\
Here are some relevant papers on this idea just for your background knowledge: \texttt{retrieval\_papers}. \\
You should generate a detailed proposal based on the given knowledge. Try to be creative. \\
The above papers are only for inspiration and you should not cite them and just make some incremental modifications. Instead, you should make sure your proposal is novel and distinct from the prior literature. \\
You should aim for projects that can potentially win best paper awards at top conferences like ACL, NeurIPS, ICLR, and CVPR. \\
The proposal should be described as: (1) Problem: State the problem statement, which should be closely related to the idea description and something that large language models cannot solve well yet. \\
(2) Existing Methods: Mention some existing benchmarks and baseline methods if there are any. \\
(3) Motivation: Explain the inspiration of the proposed method and why it would work well. \\
(4) Proposed Method: Propose your new method and describe it in detail. The proposed method should be maximally different from all existing work and baselines, and be more advanced and effective than the baselines. You should be as creative as possible in proposing new methods; we love unhinged ideas that sound crazy. This should be the most detailed section of the proposal. \\
(5) Experiment Plan: Specify the experiment steps, baselines, and evaluation metrics. \\
If \texttt{use\_few\_shot\_example} is true, you can follow these examples to get a sense of how the proposal should be formatted (but don't borrow the proposals themselves): \texttt{self.method\_proposal\_examples}. \\
You should make sure to come up with your own novel and different proposal for the specified idea: \texttt{idea}. You should try to tackle important problems that are well recognized in the field and considered challenging for current models. For example, think of novel solutions for problems with existing benchmarks and baselines. In rare cases, you can propose to tackle a new problem, but you will have to justify why it is important and how to set up proper evaluation. \\
If \texttt{use\_self\_reflection} is true, in the thinking step, you can first think of about 5 proposals and analyze the advantages and disadvantages of each of them. Your final proposal can absorb their advantages and discard their disadvantages. \\
Please write down the thinking step and your final proposal. Output the final proposal in JSON format as a dictionary, where you should generate a short proposal name (e.g., "Non-Linear Story Understanding", or "Multi-Agent Negotiation") as the key and the actual proposal description as the value (following the above format). \\
\bottomrule
\end{tabular}
\caption{Prompt for initial proposal generation.}
\label{tab:table17}
\end{table*}

\begin{table*}[ht]
\centering
\begin{tabular}{p{12cm}}
\toprule
\textbf{Adaptive Ethical Dialogue System (AEDS)} \\ \midrule
\textbf{Type:} Interactive System \\ \midrule
\textbf{Problem:} Researchers often face ethical dilemmas in real-time decision-making processes, and current large language models lack the capability to provide immediate, context-specific ethical feedback. \\ \midrule
\textbf{Existing Methods:} Existing methods include ethical guidelines and compliance checklists, but these are static and do not provide real-time, adaptive feedback. Some AI systems offer ethical scoring, but they lack the nuance and adaptability required for complex ethical decisions. \\ \midrule
\textbf{Motivation:} The inspiration for this method comes from the need for a dynamic, interactive system that can provide immediate, context-specific ethical feedback. By combining real-time dialogue with adaptive learning, we can create a system that evolves with the user's behavior and provides more relevant and accurate ethical guidance. \\ \midrule
\textbf{Proposed Method:} The Adaptive Ethical Dialogue System (AEDS) will consist of two main components: a real-time dialogue engine and an adaptive learning module. The dialogue engine will engage users in conversations to understand the context and provide immediate ethical feedback. The adaptive learning module will continuously learn from user interactions to improve the accuracy and relevance of the feedback. The system will use a combination of natural language processing, reinforcement learning, and ethical guidelines to provide nuanced and context-specific advice. Additionally, the system will include a feedback loop where users can rate the quality of the advice, which will be used to further refine the model. \\ \midrule
\textbf{Experiment Plan:} The experiment will be conducted in three phases: (1) Initial Implementation: Develop the dialogue engine and adaptive learning module. (2) Pilot Testing: Conduct pilot tests with a small group of researchers to gather initial feedback and make necessary adjustments. (3) Full-Scale Deployment: Deploy the system to a larger group of researchers and evaluate its effectiveness using metrics such as user satisfaction, accuracy of ethical advice, and improvement in ethical compliance over time. Baselines will include static ethical guidelines and existing ethical scoring systems. Evaluation metrics will include user satisfaction scores, accuracy of ethical advice, and changes in ethical compliance rates. \\ \midrule
\bottomrule
\end{tabular}
\caption{An example of initial proposal.}
\label{tab:table18}
\end{table*}

\begin{table*}[ht]
\centering
\begin{tabular}{p{12cm}}
\toprule
\textbf{Research Method Breakdown} \\ \midrule

\textbf{Role:} \\
You are an expert researcher in AI, You can break down the research method into multiple submodules. \\ \midrule

\textbf{Task:} \\
Give you an initial idea, You need to break down the method into separate modules, and you need to explain in detail the specific content, purpose, composition, and keywords of each module (keywords are used to find relevant papers on Google Scholar). Initial IDEA includes Problem, Existing Methods, Proposed Method, Experiment Plan, I will give you the initial IDEA: \{format\_plan\_json(idea)\} \\ \midrule

\textbf{Example:} \\
If self.example is provided, you can follow these examples to get a sense of how it should be formatted (but don't borrow the examples themselves). \\ \midrule

\textbf{Requirements:} \\
1. You only need to break down the method part of the idea into multiple sub-modules for detailed design. Your design cannot conflict with common sense and must be innovative, reasonable, and feasible. \\
2. You should aim for the research ideas that can potentially win best paper awards at top conferences like ACL, NeurIPS, ICLR, and CVPR. \\
3. Please output your thought process. \\
4. Please think step by step. \\ \midrule

\textbf{Input:} \\
Initial idea: \{format\_plan\_json(idea)\} \\ \midrule

\textbf{Output:} \\
Thinking: \\
\textless output your thinking process here, explain why you choose these theories to discover new ideas and why it should have a chance to win the best paper awards at top conferences.\textgreater \\
MethodSubmodulesJsonList: \\
\textless JSON\textgreater \\

\bottomrule
\end{tabular}
\caption{Prompt for method decomposition.}
\label{tab:table19}
\end{table*}

\begin{table*}[ht]
\centering
\begin{tabular}{p{12cm}}
\toprule
\textbf{Adaptive Ethical Dialogue System (AEDS) Method Decomposition} \\ \midrule
\textbf{Module 1: Real-Time Dialogue Engine} \\
   \textit{Thinking:} The real-time dialogue engine is the core of the interactive system, enabling users to communicate with the AEDS. This module needs to be robust and capable of understanding and generating human-like responses. \\
   \textit{Purpose:} To facilitate real-time interaction between the user and the AEDS, capturing the context and nuances of the user's queries. \\
   \textit{Implementation:} \\
   - \textbf{Natural Language Processing (NLP) Interface:} Utilizes advanced NLP techniques to understand and generate human-like responses. \\
   - \textbf{Dialogue Management System:} Manages the flow of conversation, ensuring coherence and relevance in responses. \\
   - \textbf{Context Capture Mechanism:} Captures the context of the conversation to provide relevant ethical feedback. \\
   \textit{Search Keywords:} real-time dialogue systems, natural language processing, interactive AI systems, context-aware dialogue \\ \midrule

\textbf{Module 2: Context Understanding Module} \\
   \textit{Thinking:} Understanding the context is crucial for providing accurate ethical advice. This module will analyze the user's input to extract relevant contextual information. \\
   \textit{Purpose:} To analyze and understand the context of the user's queries to provide accurate and relevant ethical feedback. \\
   \textit{Implementation:} \\
   - \textbf{Context Extraction Algorithms:} Algorithms to extract relevant contextual information from user input. \\
   - \textbf{Semantic Analysis Tools:} Tools to analyze the semantics of the conversation. \\
   - \textbf{Contextual Database:} A database to store and retrieve contextual information for ongoing conversations. \\
   \textit{Search Keywords:} context understanding in AI, semantic analysis, context extraction algorithms, contextual databases \\ \midrule

\textbf{Module 3: Ethical Decision-Making Engine} \\
   \textit{Thinking:} The ethical decision-making engine is responsible for generating ethical advice based on the context and ethical guidelines. This module needs to be adaptive and capable of handling complex ethical dilemmas. \\
   \textit{Purpose:} To generate ethical advice based on the context and predefined ethical guidelines. \\
   \textit{Implementation:} \\
   - \textbf{Ethical Guidelines Repository:} A repository of ethical guidelines and principles. \\
   - \textbf{Decision-Making Algorithms:} Algorithms to generate ethical advice based on the context and guidelines. \\
   - \textbf{Nuance and Adaptability Mechanisms:} Mechanisms to ensure the advice is nuanced and adaptable to complex situations. \\
   \textit{Search Keywords:} ethical decision-making in AI, adaptive ethical systems, ethical guidelines in AI, nuanced ethical advice \\ \midrule

\bottomrule
\end{tabular}
\caption{An example of method decomposition (Part 1).}
\label{tab:table20}
\end{table*}

\begin{table*}[ht]
\centering
\begin{tabular}{p{12cm}}
\toprule
\textbf{Adaptive Ethical Dialogue System (AEDS) Method Decomposition} \\ \midrule

\textbf{Module 4: Adaptive Learning Module} \\
   \textit{Thinking:} The adaptive learning module ensures that the system continuously improves based on user interactions. This module will leverage reinforcement learning to update the model. \\
   \textit{Purpose:} To continuously learn from user interactions and improve the accuracy and relevance of the ethical feedback. \\
   \textit{Implementation:} \\
   - \textbf{Reinforcement Learning Framework:} A framework to implement reinforcement learning for continuous improvement. \\
   - \textbf{User Interaction Data Collection:} Mechanisms to collect data from user interactions. \\
   - \textbf{Model Update Mechanisms:} Mechanisms to update the model based on the collected data. \\
   \textit{Search Keywords:} adaptive learning in AI, reinforcement learning, continuous improvement in AI, user interaction data \\ \midrule

\textbf{Module 5: User Feedback Loop} \\
   \textit{Thinking:} User feedback is essential for refining the system. This module will collect and analyze user feedback to improve the quality of the ethical advice. \\
   \textit{Purpose:} To collect and analyze user feedback to refine the model and improve the quality of the ethical advice. \\
   \textit{Implementation:} \\
   - \textbf{Feedback Collection Interface:} An interface for users to provide feedback on the ethical advice. \\
   - \textbf{Feedback Analysis Tools:} Tools to analyze the feedback and identify areas for improvement. \\
   - \textbf{Model Refinement Mechanisms:} Mechanisms to refine the model based on the feedback analysis. \\
   \textit{Search Keywords:} user feedback in AI, feedback-driven improvement, ethical advice refinement, user interaction feedback \\ \midrule

\textbf{Module 6: Evaluation and Metrics Module} \\
   \textit{Thinking:} Evaluating the effectiveness of the system is crucial for its success. This module will define and measure key metrics to assess the system's performance. \\
   \textit{Purpose:} To evaluate the effectiveness of the AEDS using predefined metrics and baselines. \\
   \textit{Implementation:} \\
   - \textbf{Evaluation Metrics Definition:} Define metrics such as user satisfaction, accuracy of ethical advice, and ethical compliance rates. \\
   - \textbf{Baseline Comparison:} Compare the AEDS with static ethical guidelines and existing ethical scoring systems. \\
   - \textbf{Performance Tracking Tools:} Tools to track the performance of the AEDS over time. \\
   \textit{Search Keywords:} AI system evaluation, performance metrics in AI, ethical compliance metrics, user satisfaction in AI \\ \midrule

\bottomrule
\end{tabular}
\caption{An example of method decomposition (Part 2)}
\label{tab:table21}
\end{table*}

\begin{table*}[ht]
\centering
\begin{tabular}{p{14cm}}
\toprule
\textbf{Project Proposal Generation Prompt} \\ 
\midrule

\textbf{Prompt:} \\
"You are an expert researcher in AI and your job is to expand a brief project idea into a full project proposal with detailed methodology and experiment plans so that your students can follow the steps and execute the full project. \\

The idea is: \\ 
\texttt{format\_plan\_json(idea)} \\

The target paper is the primary research study you aim to enhance or build upon through future research, serving as the central source and focus for identifying and developing the specific research idea. \\

Target paper title: \texttt{paper.title}, target paper abstract: \texttt{paper.abstract} \\

If \texttt{method\_decom\_info} exists: \\ 
You will be given a method decomposition info (possible method module design). You need to analyze whether the design of these modules is reasonable. Please learn from the good places to complete the detailed design of the final method. \\ 
Method decomposition info: \texttt{method\_decom\_info} \\

If feedback exists: \\ 
The feedback is: \texttt{feedback} \\

If new knowledge exists: \\ 
The new knowledge is: \texttt{new\_knowledge} \\

Now you should come up with the full proposal covering: \\

1. Title: A concise statement of the main research question to be used as the paper title. \\ 
2. Problem Statement: Clearly define the problem your research intends to address. Explain clearly why this problem is interesting and important. \\ 
3. Motivation: Explain why existing methods (both classic ones and recent ones) are not good enough to solve the problem, and explain the inspiration behind the new proposed method. You should also motivate why the proposed method would work better than existing baselines on the problem. \\ 
4. Proposed Method: Explain how the proposed method works, describe all the steps. Make sure every step is clearly described and feasible to implement. \\ 
5. Step-by-Step Experiment Plan: Break down every single step of the experiments, make sure every step is executable. Cover all essential details such as the datasets, models, and metrics to be used. If the project involves prompting, give example prompts for each step. \\

The experiment plan should not include any background introduction (you can skip the literature review, paper writing tips, and ethical discussion). Just give instructions on the experiments. \\

Be consistent in your methodology and experiment design, for example, if you will use black-box LLM APIs such as GPT and Claude for your experiments, then you shouldn't propose any experiments that require white-box model weights or data access and you should edit them accordingly to follow the black-box assumptions. \\

\bottomrule
\end{tabular}
\caption{Prompt for final proposal generation (Part 1).}
\label{tab:table22a}
\end{table*}

\begin{table*}[ht]
\centering
\begin{tabular}{p{14cm}}
\toprule

If \texttt{self.demo\_examples} exists: \\ 
Note that we only provide examples related to prompt work. Please refer to this format in other fields. You can follow these examples to get a sense of how the idea should be formatted (but don't bore the ideas). Below are a few examples of how the full experiment plans should look like: \\ 
\texttt{self.demo\_examples} \\

\textbf{Requirements:} \\ 
1. Consider novelty, significance, correctness, and reproducibility to ensure the high quality of the final proposal. \\ 
2. You should aim for a final proposal that can potentially win best paper awards at top conferences like ACL, NeurIPS, ICLR, and CVPR. \\ 
3. Please output your thought process. \\ 
4. Please think step by step. \\

Now please write down your final proposal in JSON format (keys should be the section names, just like the above examples). Make sure to be as detailed as possible so that a student can directly follow the plan to implement the project. \\

\bottomrule
\end{tabular}
\caption{Prompt for final proposal generation (Part 2).}
\label{tab:table22b}
\end{table*}

\begin{table*}[ht]
\centering
\begin{tabular}{p{12cm}}
\toprule
\textbf{Adaptive Ethical Dialogue System: Real-Time, Context-Specific Ethical Feedback for Researchers} \\ \midrule
\textbf{Problem Statement:} Researchers often face ethical dilemmas in real-time decision-making processes, and current large language models lack the capability to provide immediate, context-specific ethical feedback. Existing methods, such as ethical guidelines and compliance checklists, are static and do not offer real-time, adaptive feedback. \\ \midrule
\textbf{Motivation:} Existing methods, including ethical guidelines and compliance checklists, are static and do not provide real-time, adaptive feedback. Some AI systems offer ethical scoring, but they lack the nuance and adaptability required for complex ethical decisions. The inspiration for this method comes from the need for a dynamic, interactive system that can provide immediate, context-specific ethical feedback. By combining real-time dialogue with adaptive learning, we can create a system that evolves with the user's behavior and provides more relevant and accurate ethical guidance. \\ \midrule
\textbf{Proposed Method:} \\ \midrule
   \textbf{1.Real-Time Dialogue Engine:} \\
   \textit{Purpose:} To facilitate real-time interaction between the user and the AEDS, capturing the context and nuances of the user's queries. \\
   \textit{Implementation:} \\
   - \textbf{Natural Language Processing (NLP) Interface:} Utilizes advanced NLP techniques to understand and generate human-like responses. \\
   - \textbf{Dialogue Management System:} Manages the flow of conversation, ensuring coherence and relevance in responses. \\
   - \textbf{Context Capture Mechanism:} Captures the context of the conversation to provide relevant ethical feedback. \\ \midrule
   \textbf{2.Context Understanding Module:} \\
   \textit{Purpose:} To analyze and understand the context of the user's queries to provide accurate and relevant ethical feedback. \\
   \textit{Implementation:} \\
   - \textbf{Context Extraction Algorithms:} Algorithms to extract relevant contextual information from user input. \\
   - \textbf{Semantic Analysis Tools:} Tools to analyze the semantics of the conversation. \\
   - \textbf{Contextual Database:} A database to store and retrieve contextual information for ongoing conversations. \\ \midrule
   \textbf{3.Ethical Decision-Making Engine:} \\
   \textit{Purpose:} To generate ethical advice based on the context and predefined ethical guidelines. \\
   \textit{Implementation:} \\
   - \textbf{Ethical Guidelines Repository:} A repository of ethical guidelines and principles. \\
   - \textbf{Decision-Making Algorithms:} Algorithms to generate ethical advice based on the context and guidelines. \\
   - \textbf{Nuance and Adaptability Mechanisms:} Mechanisms to ensure the advice is nuanced and adaptable to complex situations. \\ \midrule
\bottomrule
\end{tabular}
\caption{An example of final proposal (Part 1).}
\label{tab:table23}
\end{table*}

\begin{table*}[ht]
\centering
\begin{tabular}{p{12cm}}
\toprule
\textbf{Step-by-Step Experiment Plan:} \\ \midrule
   \textbf{Step 1: Gather Datasets:} Collect datasets that include real-time ethical dilemmas faced by researchers. These can be sourced from academic institutions, ethical review boards, and simulated scenarios. \\
   \textbf{Step 2: Develop the Real-Time Dialogue Engine:} \\
   - \textbf{NLP Interface:} Implement an NLP interface using state-of-the-art models such as GPT-4 from the OpenAI API. \\
   - \textbf{Dialogue Management System:} Develop a system to manage the flow of conversation, ensuring coherence and relevance. \\
   - \textbf{Context Capture Mechanism:} Implement mechanisms to capture the context of the conversation. \\
   \textbf{Step 3: Implement the Context Understanding Module:} \\
   - \textbf{Context Extraction Algorithms:} Develop algorithms to extract relevant contextual information from user input. \\
   - \textbf{Semantic Analysis Tools:} Implement tools to analyze the semantics of the conversation. \\
   - \textbf{Contextual Database:} Create a database to store and retrieve contextual information. \\
   \textbf{Step 4: Develop the Ethical Decision-Making Engine:} \\
   - \textbf{Ethical Guidelines Repository:} Compile a repository of ethical guidelines and principles. \\
   - \textbf{Decision-Making Algorithms:} Develop algorithms to generate ethical advice based on the context and guidelines. \\
   - \textbf{Nuance and Adaptability Mechanisms:} Implement mechanisms to ensure the advice is nuanced and adaptable. \\
   \textbf{Step 5: Implement the Adaptive Learning Module:} \\
   - \textbf{Reinforcement Learning Framework:} Develop a framework to implement reinforcement learning. \\
   - \textbf{User Interaction Data Collection:} Create mechanisms to collect data from user interactions. \\
   - \textbf{Model Update Mechanisms:} Implement mechanisms to update the model based on the collected data. \\
   \textbf{Step 6: Develop the User Feedback Loop:} \\
   - \textbf{Feedback Collection Interface:} Create an interface for users to provide feedback. \\
   - \textbf{Feedback Analysis Tools:} Implement tools to analyze the feedback. \\
   - \textbf{Model Refinement Mechanisms:} Develop mechanisms to refine the model based on the feedback analysis. \\
   \textbf{Step 7: Implement the Evaluation and Metrics Module:} \\
   - \textbf{Evaluation Metrics Definition:} Define metrics such as user satisfaction, accuracy of ethical advice, and ethical compliance rates. \\
   - \textbf{Baseline Comparison:} Compare the AEDS with static ethical guidelines and existing ethical scoring systems. \\
   - \textbf{Performance Tracking Tools:} Implement tools to track the performance of the AEDS over time. \\
   \textbf{Step 8: Pilot Testing:} Conduct pilot tests with a small group of researchers to gather initial feedback and make necessary adjustments. \\
   \textbf{Step 9: Full-Scale Deployment:} Deploy the system to a larger group of researchers and evaluate its effectiveness using the defined metrics. \\ \midrule
\bottomrule
\end{tabular}
\caption{An example of final proposal (Part 2).}
\label{tab:table24}
\end{table*}

\begin{table*}[ht]
\centering
\begin{tabular}{p{12cm}}
\toprule
\textbf{Prompt for Research Search Plan} \\
\midrule
\textbf{Role:} You are an expert researcher in AI. You can think out of the box and develop a detailed paper search plan for a given research idea. \\
\textbf{Skill:} Given a research idea, analyze which fields' papers should be searched in order to collect comprehensive information and new knowledge for further study and finding new ideas. Please provide the thinking process and search keywords. \\
\midrule
\textbf{Example:} \{self.few\_shot\_example\} \\
\textbf{Research Plan Format:} \{self.plan\_json\_format\}\\
\midrule
\textbf{Requirement:} \\
\begin{enumerate}
    \item The search plan needs to be developed around the given research idea.
    \item Please output your thought process.
    \item Please think step by step.
\end{enumerate} \\
\textbf{Input:} Research idea: \{idea\_info\} \\
\textbf{Output:} \\
\textbf{Thinking:} \\
\textbf{Search Plan Output:} \\
\texttt{<JSON>} \\
\bottomrule
\end{tabular}
\caption{Prompt for search plan generation.}
\label{tab:table25}
\end{table*}

\begin{table*}[ht]
\centering
\begin{tabular}{p{12cm}}
\toprule
\textbf{Research Idea and Search Plan} \\
\midrule
\textbf{Research Idea:} Exploring the long-term impact of LLM-generated research ideas on scientific progress and innovation. \\
\textbf{Keywords:} long-term impact, scientific progress, innovation, LLM-generated research ideas \\
\textbf{Thinking:} Understanding the long-term effects can guide future development and application of LLM in research. \\
\midrule
\textbf{Search Plan Output:} To explore the long-term impact of LLM-generated research ideas on scientific progress and innovation, we need to collect relevant literature from multiple disciplines and fields. Here are the suggested query areas, thought processes, and keywords for each area: \\
\midrule
\textbf{1. How LLM or ChatGPT Generates Research Ideas:} \\
\textbf{Thinking:} Understanding the development of LLM technology itself, especially how LLM generates research ideas. This involves machine learning models, algorithms, and their applications in generating new ideas. \\
\textbf{Keywords:} LLM research idea generation, LLM inspiration generation, LLM creativity, OpenLLM, GPT-4, Ideate \\
\midrule
\textbf{2. Philosophy of Science and History of Science:} \\
\textbf{Thinking:} Studying the impact of LLM on scientific progress requires examining the nature and process of scientific development from a philosophical and historical perspective. This helps us understand how LLM may change the way of scientific exploration. \\
\textbf{Keywords:} philosophy of science, history of scientific innovation, scientific methodology, LLM in scientific history \\
\midrule
\textbf{3. Sociology and Sociology of Science:} \\
\textbf{Thinking:} The sociological perspective can help us understand how LLM-generated research ideas affect the structure and dynamics of the scientific community, and how these changes affect scientific progress. \\
\textbf{Keywords:} sociology of science, scientific community, LLM impact on scientific sociology, social dynamics of innovation \\
\midrule
\textbf{4. Policy Research and Science and Technology Policy:} \\
\textbf{Thinking:} Policy research can provide insights on how to guide the application of LLM in scientific research through policies, and how these policies affect scientific progress and innovation. \\
\textbf{Keywords:} science policy, LLM policy, long-term policy impact, innovation policy \\
\bottomrule
\end{tabular}
\caption{An example of search plan.}
\label{tab:table26}
\end{table*}

\begin{table*}[ht]
\centering
\begin{tabular}{p{12cm}}
\toprule
\textbf{Research Idea and Search Plan} \\
\midrule
\textbf{Research Idea:} Seed Idea 0 \\
\textbf{Keywords:} real-time data integration, recent publications tracking, innovative research ideas \\
\textbf{Thinking:} The target paper's reliance on existing literature may limit the generation of truly novel ideas. By incorporating real-time data sources, such as ongoing research developments or recent publications, the ResearchAgent can generate more innovative and timely research ideas. \\
\midrule
\textbf{Plan Thinking List:} \\
First, we need to understand the technical methods for integrating real-time data feeds into a knowledge base. This involves exploring various data integration frameworks, tools, and techniques. \\
Next, we need to explore methods for tracking recent publications and ongoing research developments. This includes automated literature review tools, databases, and alert systems. \\
We should also explore strategies and tools for generating innovative research ideas. This includes AI-driven idea generation, creativity tools, and methodologies for fostering innovation. \\
We need to understand how to manage and update a knowledge base effectively, especially when integrating real-time data and recent publications. This involves exploring knowledge management systems and update protocols. \\
Finally, we should explore the impact of real-time data integration on research innovation. This involves studying case studies, theoretical frameworks, and empirical evidence on how real-time data influences research outcomes. \\
\midrule
\textbf{New Knowledge List:} \\
\textbf{Title:} Learnings from Data Integration for Augmented Language Models \\
\textbf{Abstract:} One of the limitations of large language models is that they do not have access to up-to-date, proprietary or personal data. \\

\textbf{Title:} SciNews: From Scholarly Complexities to Public Narratives -- A Dataset for Scientific News Report Generation \\
\textbf{Abstract:} Scientific news reports serve as a bridge, adeptly translating complex research articles into reports that resonate with the broader public. \\

\midrule
\textbf{Expanded Ideas:} \\
\textbf{Idea:} Dynamic Knowledge Graph Integration \\
\textbf{Thinking:} Integrating dynamic knowledge graphs can significantly enhance the ResearchAgent’s ability to connect disparate pieces of information dynamically. \\
\textbf{Keywords:} dynamic knowledge graphs, real-time data integration, knowledge base management \\

\textbf{Idea:} Context-Aware Research Idea Generation \\
\textbf{Thinking:} Context-aware research idea generation can generate ideas based on the context of ongoing research trends and developments. \\
\textbf{Keywords:} context-aware idea generation, automated literature review, real-time research updates \\

\textbf{Idea:} Real-Time Peer Review Feedback Integration \\
\textbf{Thinking:} Real-time peer review feedback integration can incorporate real-time feedback from peer reviews to refine research ideas iteratively. \\
\textbf{Keywords:} real-time peer review, feedback integration, human preference-aligned models \\
\bottomrule

\end{tabular}
\caption{An example of search result.}
\label{tab:table27}
\end{table*}

\end{document}